\documentclass[12pt]{iopart}

\usepackage{subfigure}
\usepackage{algorithm}
\usepackage{algorithmicx}
\usepackage{algpseudocode}
\usepackage{bm}
\expandafter\let\csname equation*\endcsname\relax
\expandafter\let\csname endequation*\endcsname\relax
\usepackage{amsmath}
\usepackage{amsthm}
\usepackage{amssymb}
\usepackage{empheq}
\newtheorem{theorem}{Theorem}
\newtheorem*{remark}{Remark}
\newtheorem*{pproof}{\bf{Proof}}
\usepackage{cite}

\begin{document}
\title[]{Mixed geometry information regularization for image multiplicative denoising}
\author{Shengkun Yang$^{1}$, Zhichang Guo$^{1}$, Jia Li$^{1,}$\footnote[7]{Author to whom any correspondence should be addressed.}, Fanghui Song$^1$, Wenjuan Yao$^{1}$} 
\address{1 School of Mathematics, Harbin Institute of Technology, Harbin 150001, China}
\ead{jli@hit.edu.cn}
\vspace{10pt}
\begin{indented}
\item[]29 April 2024
\end{indented}

\begin{abstract}
This paper focuses on solving the multiplicative gamma denoising problem via a variation model. Variation-based regularization models have been extensively employed in a variety of inverse problem tasks in image processing. However, sufficient geometric priors and efficient algorithms are still very difficult problems in the model design process. To overcome these issues, in this paper we propose a mixed geometry information model, incorporating area term and curvature term as prior knowledge. In addition to its ability to effectively remove multiplicative noise, our model is able to preserve edges and prevent staircasing effects. Meanwhile, to address the challenges stemming from the nonlinearity and non-convexity inherent in higher-order regularization, we propose the efficient additive operator splitting algorithm (AOS) and scalar auxiliary variable algorithm (SAV). The unconditional stability possessed by these algorithms enables us to use large time step. And the SAV method shows higher computational accuracy in our model. We employ the second-order SAV algorithm to further speed up the calculation while maintaining accuracy. We demonstrate the effectiveness and efficiency of the model and algorithms by a lot of numerical experiments, where the model we proposed has better features texture-preserving properties without generating any false information.
\end{abstract}

%
\vspace{0.1pc}
\hspace{1.65cm}{\it Keywords}: multiplicative denoising, mixed geometry information, \par
\hspace{1.65cm}additive operator splitting, scalar auxiliary variable
%
%
%
%

\section{Introduction}
\label{intro}
Multiplicative noise, commonly referred to as speckle noise, is prevalent in the synthetic aperture radar (SAR) images \cite{fitch2012synthetic,oliver2004understanding}, laser images \cite{goodman1975statistical}, ultrasound images \cite{wagner1983statistics}, and tomographic images \cite{ollinger1997positron}. It destroys almost all the original information, hence it is necessitating urgent attention to perform multiplicative noise removal in real imaging systems. For
a mathematical description of such noise, we consider a clean image $u :\Omega \rightarrow \mathbb{R}$, where $\Omega$ is a bounded open set in $\mathbb{R}^2$ with a Lipschitz boundary. Let $\eta$ represent the multiplicative noise, then the pollution process of noisy image $f$ can be expressed mathematically as follows
\begin{equation}
	f=u \eta.
\end{equation}
In this study, our focus is on the consideration that $\eta$ follows the gamma distribution \cite{bamler2000principles}, which is commonly encountered in SAR images. Specifically, $\eta$ obeys the mean value of $1$ and the variance of $1/L$. The probability density function is formally expressed as
\begin{equation}
	p(\eta)=\frac{L^L \eta^{L-1}}{\Gamma(L)} \mathrm{e}^{-L \eta} \mathbf{1}_{\{\eta \geq 0\}},
\end{equation}
where $\mathbf{1}_{\{\eta \geq 0\}}$ denotes the characteristic function of the set $\{\eta \geq 0\}$. Different from the additive noise, multiplicative noise exhibits coherence and non-Gaussian properties \cite{feng2021models}, making traditional additive denoising models fail and posing more challenges.

In the literature, various multiplicative denoising methods have been proposed, including variational methods \cite{aubert2008variational,shi2008nonlinear,jin2010analysis,ullah2017new}, partial differential equation (PDE) methods \cite{zhou2014doubly,yao2019multiplicative}, non-local filtering methods \cite{baraha2022systematic,teuber2012new,deledalle2014nl,penna2019sar}, and so on. 
Among these, variation-based models renowned for their adeptness in incorporating prior information and possessing stringent theoretical guarantees, and have attracted significant attentions \cite{osher2003multiplicative,aubert2008variational,shi2008nonlinear,huang2009new,zhao2014new}.
A classic model within this category is the Rudin, Lions, and Osher (RLO) model \cite{osher2003multiplicative}. 
Notably, RLO utilized total variation (TV) as a regularizer, allowing for the existence of discontinuous solutions. 
It can effectively remove internal noise while preserving boundaries. 
However, the model solely considers the statistical information of mean and variance, thereby constraining its denoising capabilities. To address this limitation, Aubert and Aujol \cite{aubert2008variational} proposed the AA model, which combined the TV regularizer with a modified fidelity term ${\rm {log}}u+f/u$ derived from maximum a posteriori estimation. 

While the classic TV regularizer and its variants can effectively remove noise, they often exhibit defects, such as the staircasing effects, corners loss and contrast reduction \cite{chan2005recent,strong2003edge,zhu2012image,MR3152107}. In response to these limitations, numerous curvature-based models have been proposed. Curvature serves as a crucial tool for characterizing curves and surfaces in image processing. Models incorporating such priors can enforce continuity constraints on the solution space, allowing for the preservation of jumps and mitigation of the aforementioned deficiencies. Meanwhile, as models containing a type of geometric information, the curvature-based models also exhibit good performance in protecting image geometric structure, such as corners \cite{khan2015higher,gun2017improved,ren2018optimization,xu2018variational,zhang2022image}.

Besides employing curvature-based models, denoising can also be achieved by minimizing the area of image surface \cite{sapiro1993invariant,yezzi1998modified,kimmel2000images,qiu2011edge}. Sapiro et al. \cite{sapiro1993invariant} first minimize the area of surfaces to penalize the roughness of the solution, which can avoid smoothing the edges while denoising. Yezzi \cite{yezzi1998modified} proposed a modified momentum model from the non-variational perspective, and the corresponding model can flatten out oscillations while protecting the edge.  

In conclusion, compared with the TV-norm based models, the geometry-based models can effectively remove the multiplicative noise while protecting image features. However, it can be seen that both types of models suffers from some limitations. For example, 
the model based only on surface area has the advantage of preserving edges and contours but also brings problems such as contrast reduction and staircasing effects. 
The model based solely on curvature can avoid the above problems, but it also brings the disadvantage of being unable to protect the edges of objects and making the boundaries too smooth \cite{lysaker2003noise}. 

Then it is natural to consider a model that mixes geometric information to compensate for their shortcomings. Here we use a multiplicative structure similar to Euler's elastica to balance their effects. We take the minimal surface area term as the main denoising term, which has excellent edge protection ability and low computational cost. However, it is accompanied by unnatural artifacts such as the staircasing effects. We combine the curvature term with it to alleviate this phenomenon. In addition, we incorporate the gray level indicator to perform adaptive multiplicative denoising. 

To sum up, the model we proposed by considering curvature and surface area, not only possesses strong denoising capabilities but also preserves the geometric information of images, such as texture, edges, and corners. However, the integration of curvature and surface results in a high-order nonlinear functional model, which poses challenge in numerical solving. Frequently, related models employ rapid algorithms like the augmented Lagrangian method (ALM) and primal-dual method to expedite computations. Nevertheless, such approach introduces the issue of manually selecting numerous multipliers, resulting in additional parameter adjustment burden \cite{tai2011fast,tai2013fast,duan2013fast,zhu2013augmented,Calatroni_2020}. Addressing these challenges necessitates the development of an efficient and accurate numerical algorithm for minimizing the energy functional. To maintain equivalence during algorithmic iterations, we first employ the additive operator splitting (AOS) algorithm to solve the gradient flow. The AOS algorithm is a classic algorithm used in image processing and has unconditional stability. However, when applied to our model, it sometimes leads to computational inaccuracies. It is worth mentioning that in recent years, Shen et al. \cite{shen2018scalar} have proposed the scalar auxiliary variable (SAV) method to speed up the solving process of gradient flow problems. It is also unconditionally stable and capable of iteratively minimizing the original energy functional without necessitating modification or relaxation of the initial problem. Therefore the first-order and second-order SAV algorithm is used to calculate our model.

The main contributions of our work mainly lie in the following three aspects:

\textbullet We propose an adaptive multiplicative denoising model incorporating mixed geometry information. We mix the area term and the curvature term to protect edges and mitigate staircasing effects respectively. Furthermore, tailored to the coherence characteristics of multiplicative noise, we introduce gray level indicator to perform stronger denoising on areas with high signal intensity.

\textbullet To overcome the numerical challenges inherent in solving nonlinear models, we utilize the AOS and SAV algorithms to expedite the conventional gradient flow method. These algorithms exhibit unconditional stability and unconditional energy stability, respectively. Compared with the AOS algorithm, the SAV algorithm has a better ability to maintain accuracy in our model.

\textbullet Adequate numerical experiments demonstrate that the proposed model effectively removes noise while preserving the geometric features of the original image, including corners and edges.

The rest of this paper is organized as follows. We propose a model based on mixed geometry information regularization in Sect. 2. In Sect. 3 we derive the gradient flow corresponding to the mixed geometric information model. In Sect. 4, the unconditionally stable algorithms based on AOS and SAV are proposed. Sect. 5 is devoted to numerical experiment. In Sect. 6, we summarize the whole paper.

\section{Adaptive Minimization Model Based on Mixed Geometry Information}
\label{sec:1}
\subsection{The Comparison between Single Geometric Information and Mixed Geometric Information}
\label{sec:1subsec:1}
Among traditional variational denoising methods, regularization models based on geometric information exhibit superior performance. Models with TV regularization or minimal surface regularization have the advantage of edge protection, but these models bring bad experimental effects such as loss of image contrast \cite{zhu2012image}. At the same time, the curvature-based model can effectively maintain contrast, but the edge protection ability is reduced and the computational overhead increases several times \cite{ren2018optimization,wang2022efficient}. We can find that models based on single geometric information enjoy some advantages but also have disadvantages. Then a natural idea is that we can design a mixed geometric model that can combine their advantages and compensate for each other's shortcomings.

\begin{figure}[h]
	\begin{minipage}{\textwidth}
		\centering
		\subfigcapskip=-15pt 
		\subfigure[Minisurface]{
			\includegraphics[width=0.33\textwidth]{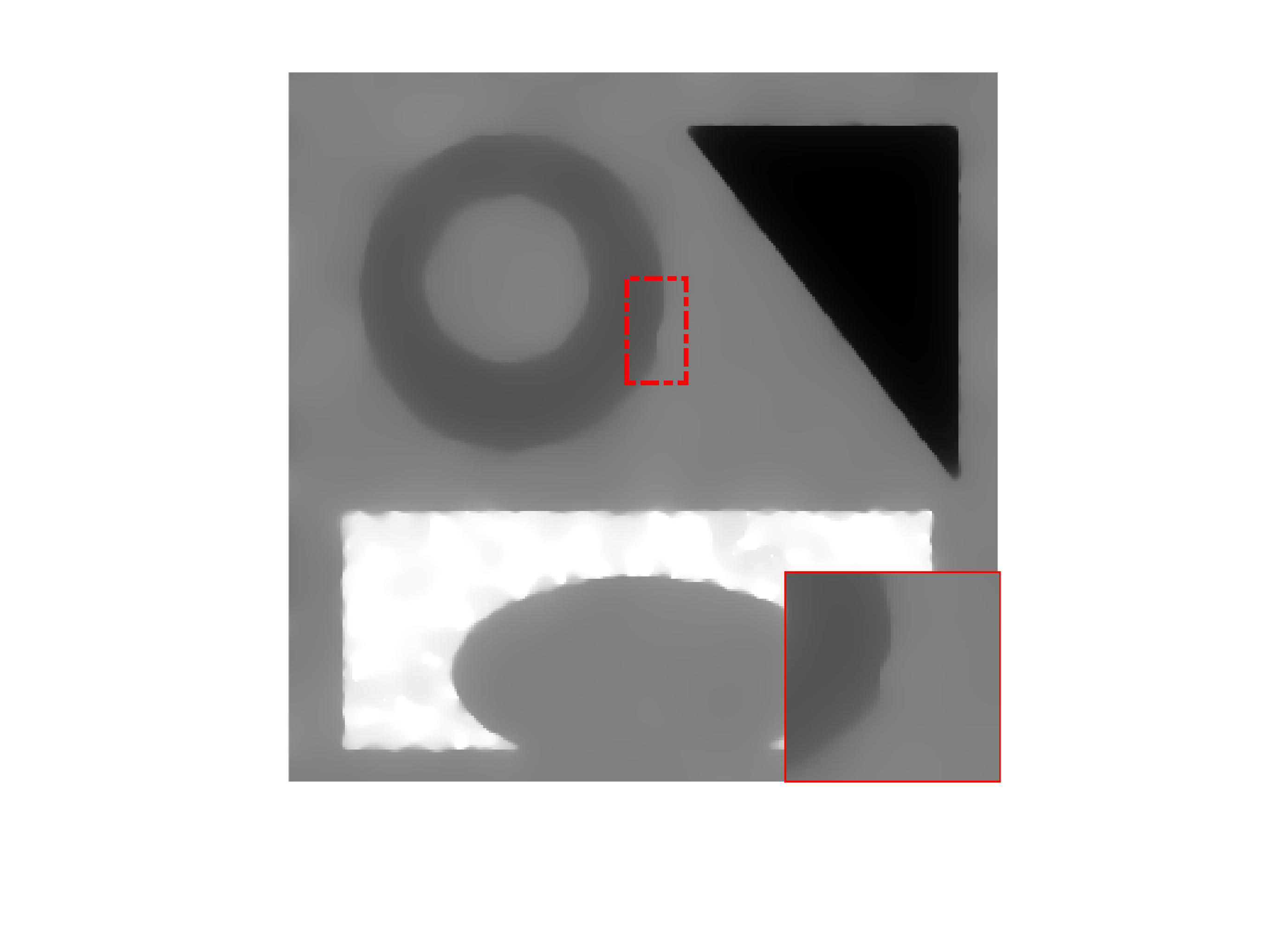}
		}\hspace{-7.5mm}
		\subfigure[Curvature]{
			\includegraphics[width=0.33\textwidth]{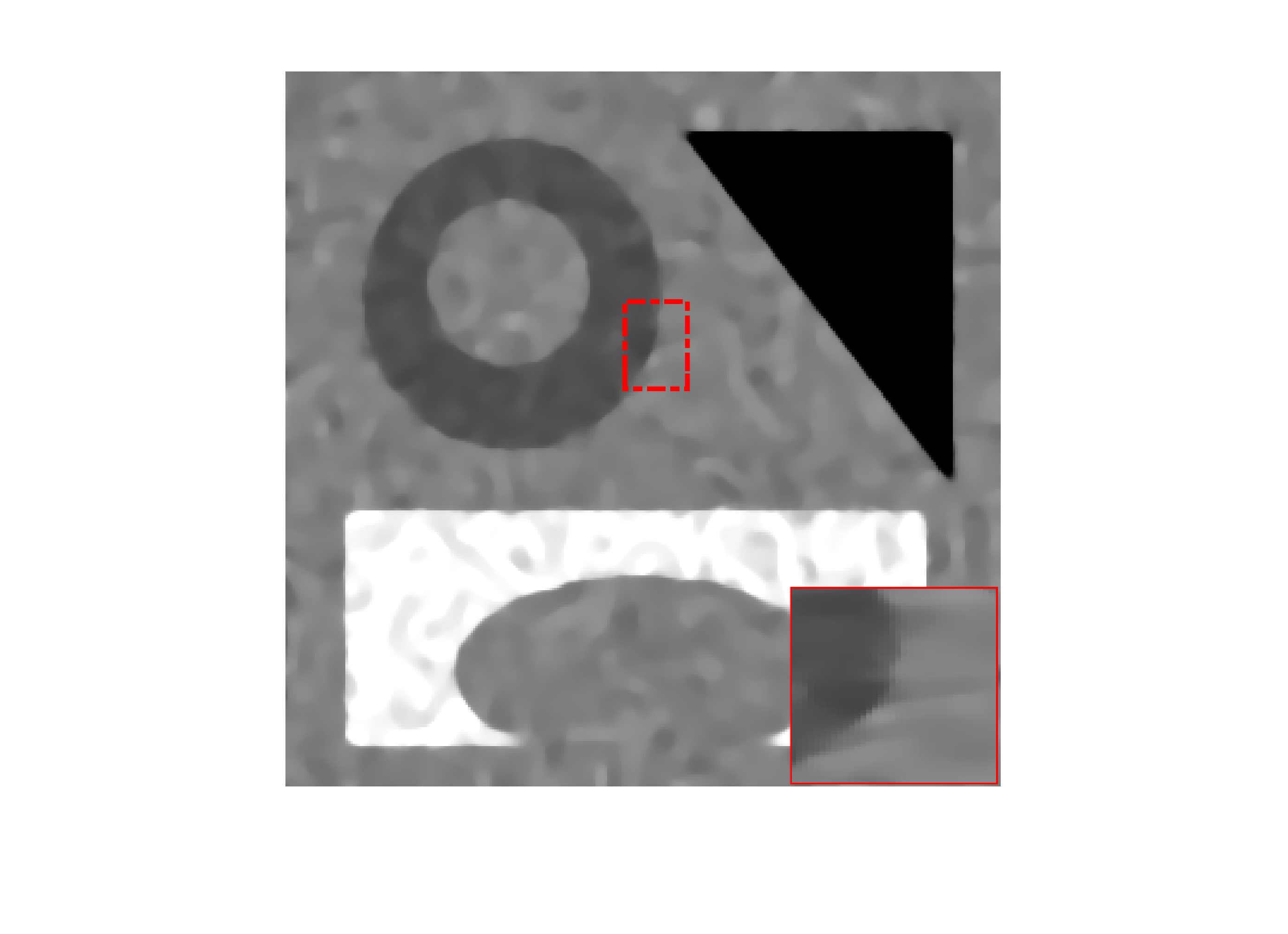}
		}\hspace{-7.5mm}
		\subfigure[Mixed geometric]{
			\includegraphics[width=0.33\textwidth]{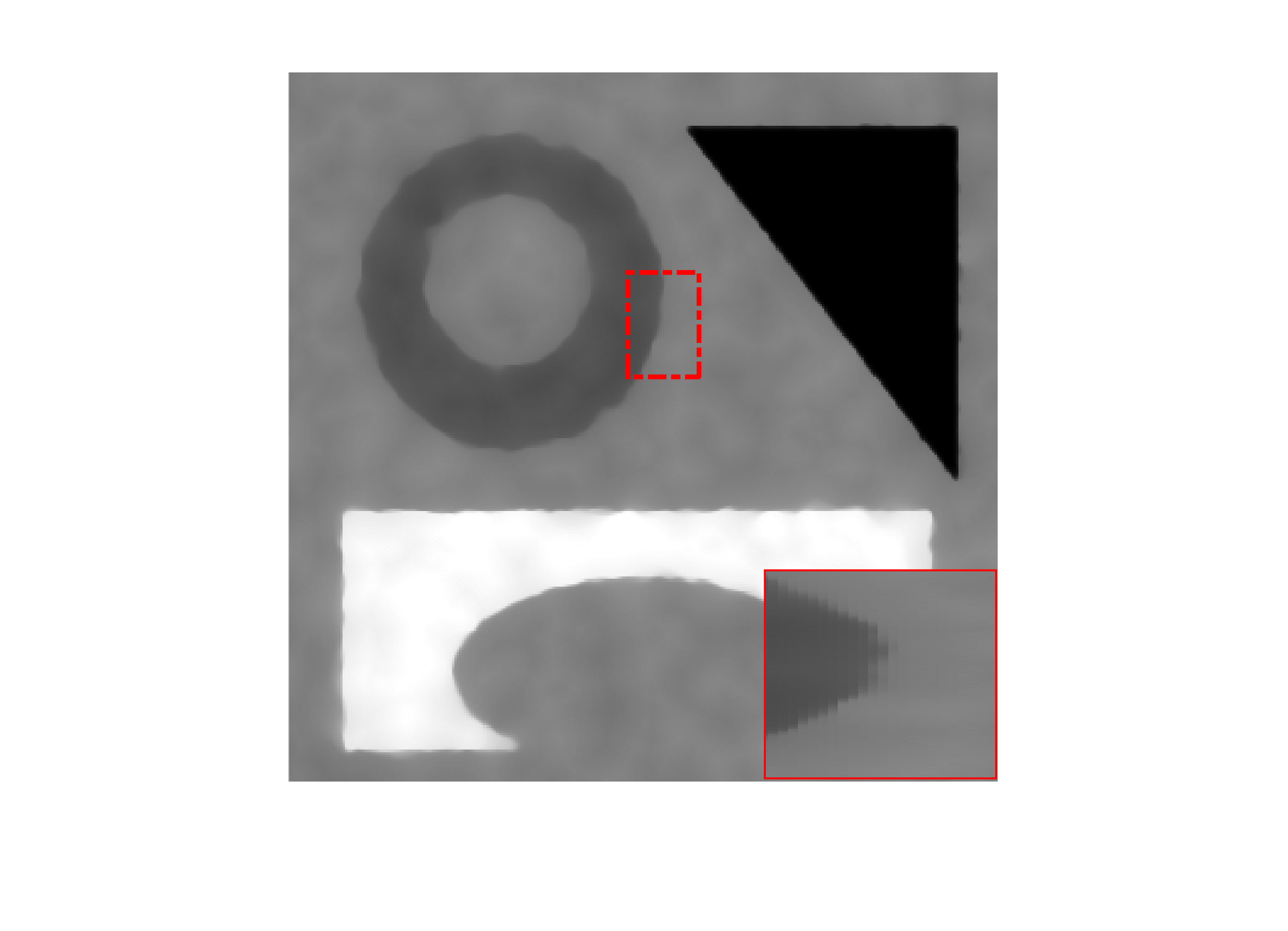}
		}
	\end{minipage}
	\caption{Denoising results of single geometric information and mixed geometric information. a minmimal surface (PSNR/SSIM:27.2070/0.8728); b curvature (PSNR/SSIM:27.2534/0.8476); c mixed geometric information (PSNR/SSIM:28.4377/0.8938)}
	\label{ad_mixgeo}
\end{figure}

Suppose the image are degraded by the multiplicative gamma noise pollution level $L=4$. As displayed in Fig. \ref{ad_mixgeo}, during the multiplicative denoising process, the minimal surface model has lower image contrast than the model based on curvature. In terms of visual effects, curvature-based model can protect image details better but over-smooths the edges. To further improve their denoising performance, we use a mixed geometric information model. As shown in the results in Fig. \ref{ad_mixgeo}, the mixed geometric information model has the best edge structure protection capabilities and comprehensive results.
\subsection{Gray Level Indicator}
\label{sec:1subsec:2}

\begin{figure}[h]
	\centering
	\subfigcapskip=-10pt
	\subfigure[Residual]{
		\includegraphics[width=0.45\textwidth]{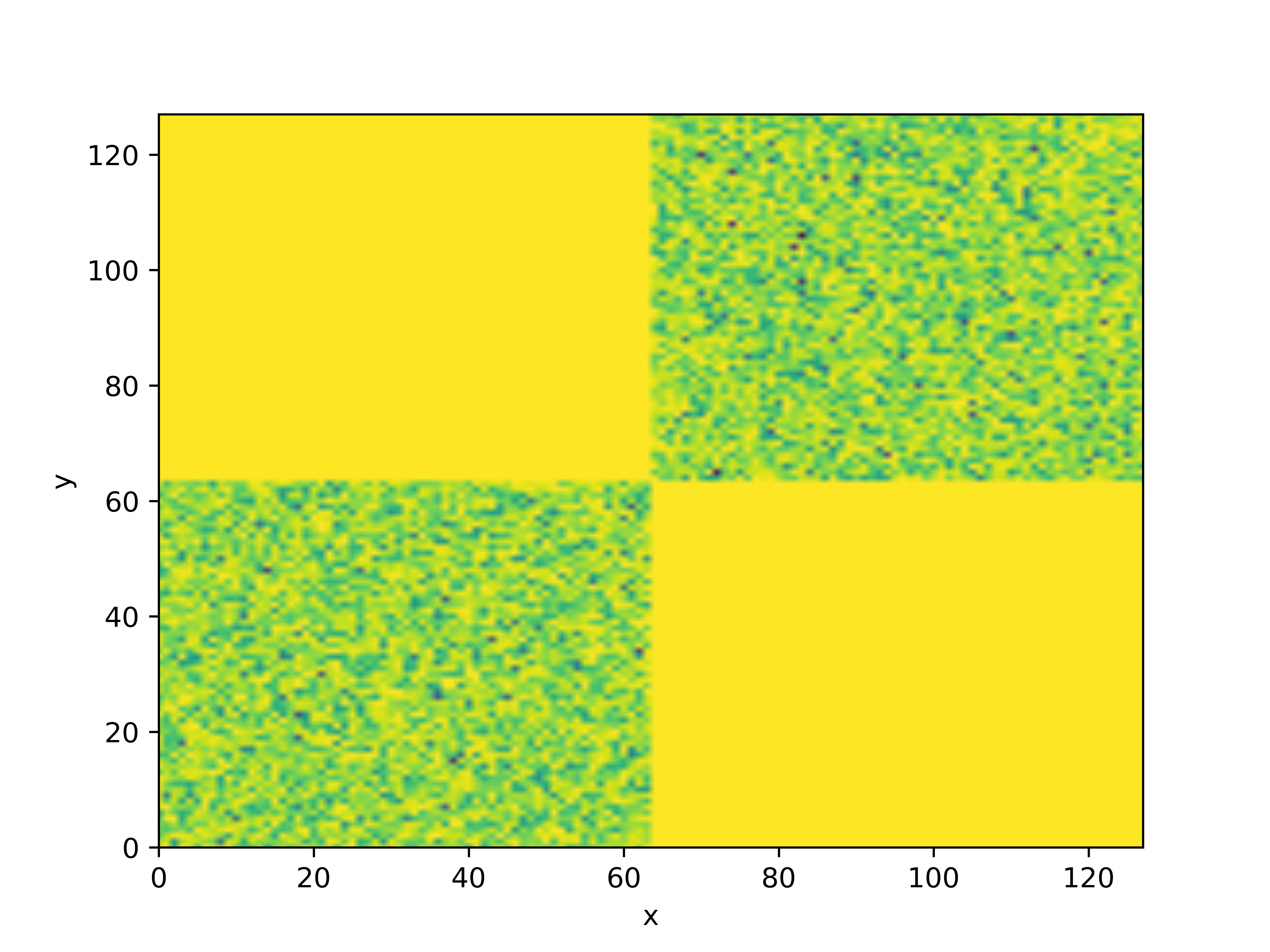}
	}
	\subfigure[Indicator]{
		\includegraphics[width=0.45\textwidth]{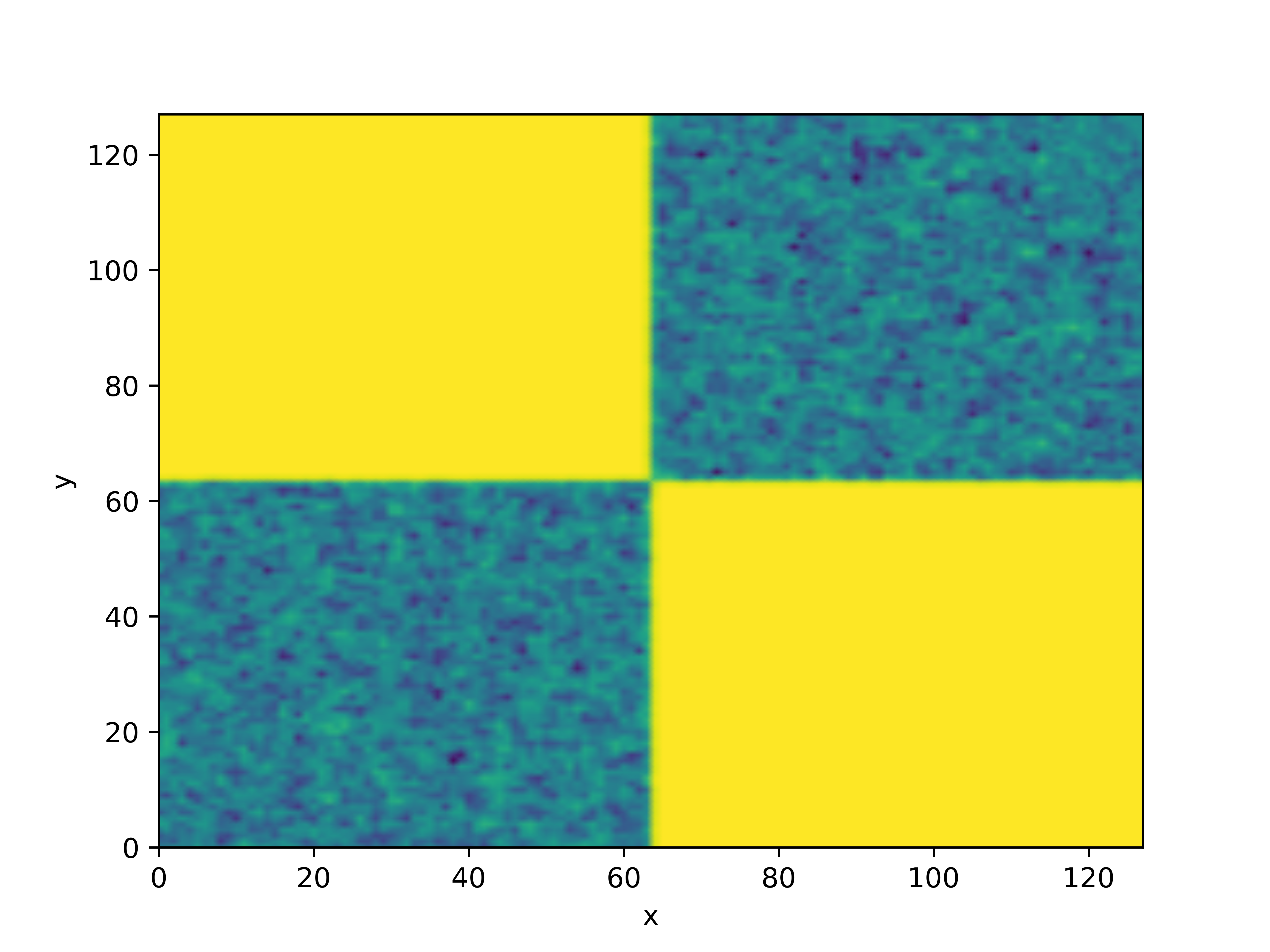}
	}
	\caption{Noise pollution level and gray level indicator}
	\label{alp}
\end{figure}

Unlike additive noise, the extent of corruption caused by multiplicative noise is linked to the signal strength of its clean image. Based on the above analysis, we design gray level indicator for adaptive denoising. Regions with higher original signal strength indicate a greater degree of pollution, necessitating stronger denoising in such regions. However, multiplicative noise severely destroys almost all information in the clean image, making it challenging to directly obtain a corrupted image. Therefore, to achieve our objective, we use a modified gray level indicator. For the image $u(x,y)$, the gray level indicator $\alpha$ can be expressed as
\begin{equation}
	\alpha(x)=\left(\frac{G_{\sigma}* u}{M}\right)^p,
\end{equation}
where $G_{\sigma}$ is the two-dimensional convolution kernel with standard deviation $\sigma$, $M=\sup_{x\in\Omega}\left(G_{\sigma}* u\left(x\right)\right)$. $\alpha$ utilizes a Gaussian convolution kernel to pre-smooth the noisy image and approximate the clean image. This operation makes the indicator less sensitive to noise of scale smaller than $\sigma$. Normalization allows the indicator to be regarded as a weight function. The higher the signal strength, the more attention it warrants, and active minimization of the surface should be prioritized. As depicted in the Fig. \ref{alp}, the indicator effectively characterizes the pollution degree of the noisy image and accurately captures the spatial variation of the image.
\subsection{Proposed Model}
\label{sec:1subsec:3}

Assuming that the gray image $u(x,y)$ is defined on the domain $\Omega$, and is considered as a regular surface in three-dimensional space. Encouraged by the success of mean curvature in handling additive noise, we incorporate the mean curvature of surfaces into the our proposed model. We reconsider the image surface as the zero level set of the level set function $\phi$ \cite{MR1862194}, $\phi(x,y,z) = u(x,y) - z$. Therefore, its unit outer normal is
\begin{equation}
	N=\frac{\nabla \phi}{|\nabla \phi|}=(n_1,n_2,n_3).
\end{equation}
Further, the mean curvature of a surface is defined as the divergence of the unit normal
\begin{equation}
	\kappa=\nabla \cdot N=\nabla \cdot \frac{\nabla \phi}{|\nabla \phi|}=\frac{\partial n_1}{\partial x}+\frac{\partial n_2}{\partial y}+\frac{\partial n_3}{\partial z}.
\end{equation}
Clearly, the smoothness of the image can be effectively represented by the curvature. Through such a transformation with the level set method, the value of $\phi$ can be used to rapidly and accurately calculate the mean curvature using methods such as the finite difference method.

Considering that the mean curvature regularization of the surface is used in the model, we regularize the surface area rather than directly limiting the length of the level set curve. This also has the advantage that it avoids the singularity caused by division by zero. Specifically, constraining all contours and regularizing the elastic energy of the level set function, we get the following regularization term
\begin{equation}
	\begin{aligned}
		&\int_{R}\int_{\phi=0}(a+b\kappa^2)\mathrm{d}s\mathrm{d}z\\
		=&\int_{R}\int_{\phi=0}(a+b\kappa^2)\sqrt{1+|\nabla u|^2}\mathrm{d}t\mathrm{d}z\\
		=&\int_{\Omega}(a+b\kappa^2)\sqrt{1+|\nabla u|^2}\mathrm{d}x.
	\end{aligned}
\end{equation}
Further, by combining the gray level indicator and fidelity term designed for the multiplicative noise process, we propose a model $E\left(u\right)$ that mixes area and curvature geometric information
\begin{equation}
	\int_{\Omega}\left(\alpha(u)+b\left(\nabla \cdot \left(\frac{\nabla u}{\sqrt{1+|\nabla u|^2}}\right)\right)^2\right) \sqrt{1+|\nabla u|^2} \mathrm{~d} x+\eta \int_{\Omega}(u-f \log u) \mathrm{d} x,
	\label{functional}
\end{equation}
where $\alpha(u)$ is the gray level indicator, and $b$ and $\lambda$ are positive constants utilized to balance the fidelity term and the regularization term. To better address the characteristics of multiplicative denoising and achieve the regional adaptive effect, we employ the gray level indicator $\alpha$ from \cite{zhou2014doubly}. To further illustrate the rationality of the model, we will further introduce the degradation model of the model, each of which is a well-known classic denoising model.

\subsection{Degenerate Case of the Model}
\label{sec:1subsec:4}
In order to further illustrate the advantages of mixing area terms with curvature terms, we conducted the following experiments. In particular, the free parameters and adaptive part of the model can degenerate the model into many simpler forms despite the high complexity of the model. (1) $\alpha(u) \neq 0$ and is a constant, $b \neq 0$, degenerate into regularization resembling Euler's elastica; (2) $\alpha(u) \neq 0$ and is a constant, $b = 0$, degenerate into minimal surface regularization; (3) $\alpha(u) $ is not a constant, $b = 0$, degenerate into adaptive minimal surface regularization; (4) $\alpha(u) $ is not a constant, $b \neq 0$, it is a multiplicative deduction based on the mixed geometry information of the surface. We next analyze various visual denoising effects of the model.

\begin{figure}[h]
	\centering
	\begin{minipage}{\textwidth}
		\centering
		\subfigcapskip=-15pt 
		\subfigure[Clean]{
			\includegraphics[width=0.33\textwidth]{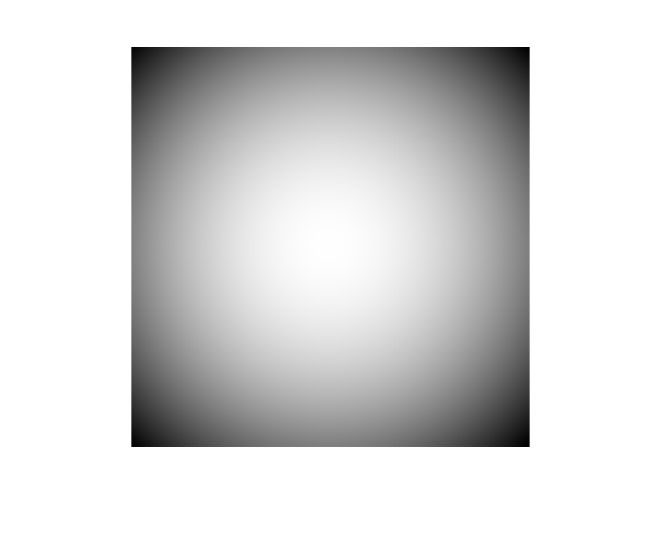}
		}\hspace{-10.5mm}
		\subfigure[Minisurface]{
			\includegraphics[width=0.33\textwidth]{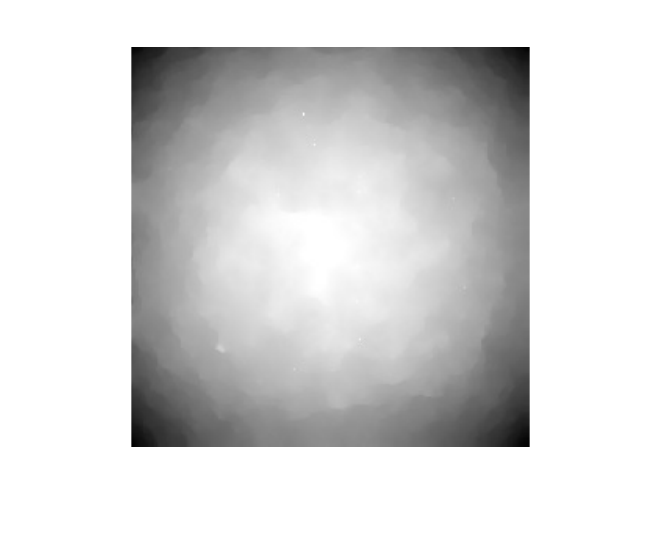}
		}\hspace{-10.5mm}
		\subfigure[Elastica]{
			\includegraphics[width=0.33\textwidth]{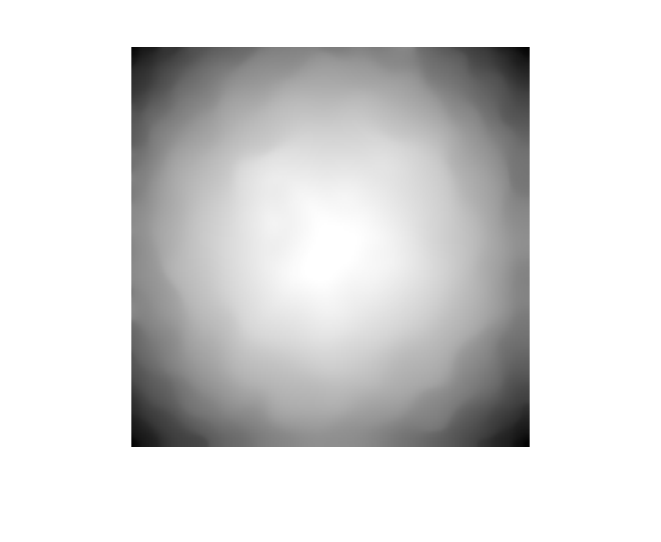}
		}
	\end{minipage}
	\centering
	\begin{minipage}{\textwidth}
		\centering
		\subfigcapskip=-15pt 
		\subfigure[$\alpha=0.5$,$b\neq0$]{
			\includegraphics[width=0.33\textwidth]{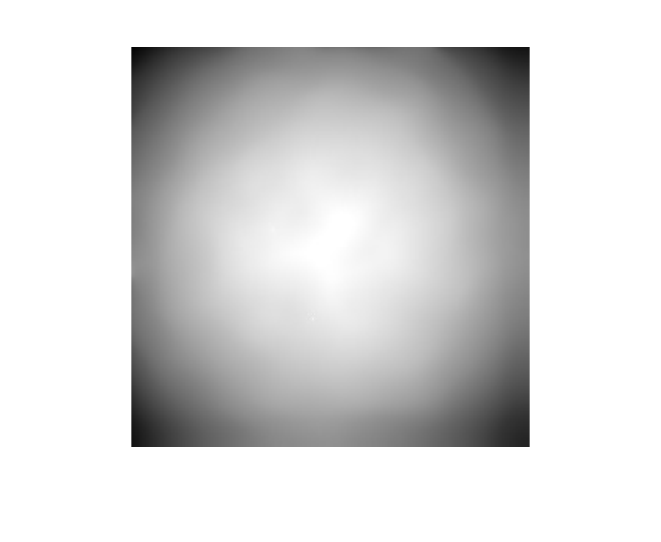}
		}\hspace{-10.5mm}
		\subfigure[$\alpha=\alpha\left(u\right)$,$b=0$]{
			\includegraphics[width=0.33\textwidth]{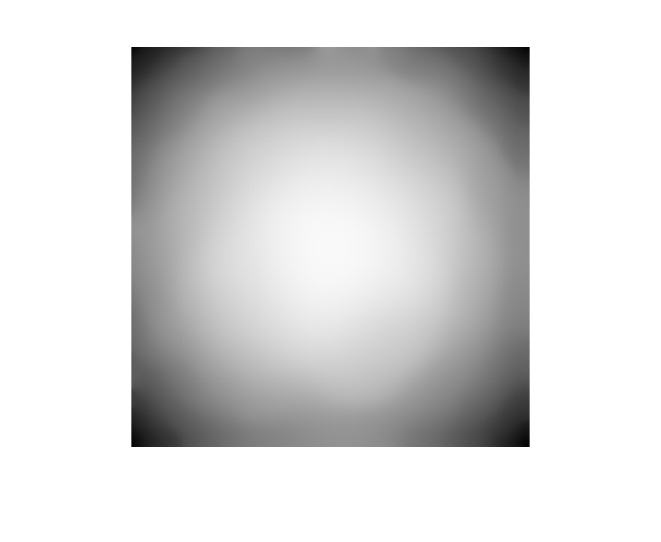}
		}\hspace{-10.5mm}
		\subfigure[$\alpha=\alpha\left(u\right)$,$b\neq0$]{
			\includegraphics[width=0.33\textwidth]{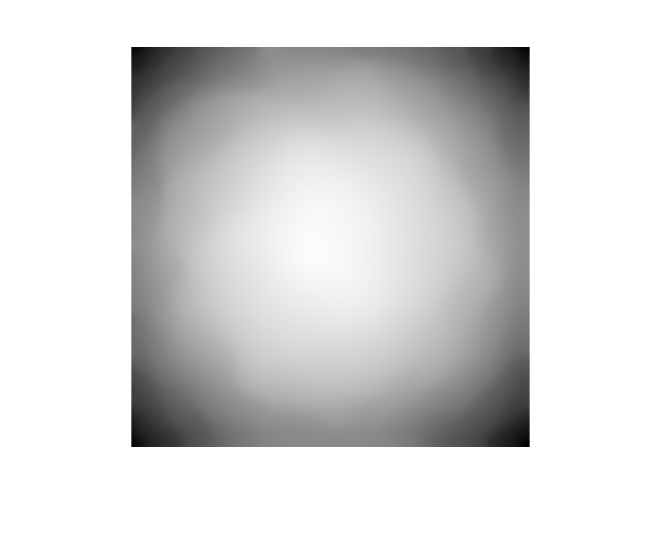}
		}
	\end{minipage}
	\caption{Model Degradation Analysis}
	\label{tuihua}
\end{figure}
In the experiment, the noise parameter $ L = 4 $ was initially selected. The traditional minimal surface model exhibits a conspicuous phenomenon of fragmentation, which leads to more false boundaries in the result, which is not conducive to the subsequent analysis. Upon employing the gray level indicator, Fig. \ref{tuihua}d illustrates a significant reduction in the staircasing effects of the minimal surface model, and Fig. \ref{tuihua}e shows a smoother result. However, a ``fish scale phenomeno'' emerges, accompanied by noticeable jagged edges at the shape's periphery. In contrast to previous models, our proposed model effectively restores image details and successfully avoids the staircasing effects. The amalgamation of area and curvature geometric information yields mutual benefits, surpassing the denoising capabilities of each individual geometric feature in multiplicative denoising.

\section{Fourth-Order Euler-Lagrange Equation of mixed geometry information model}
\label{sec:3subsec:1}
In view of the gradient flow approach, we discretize the Euler-Lagrange equation to solve the functional minimization problem. This kind of continuous approach can accurately depict the geometric characteristics of the image without causing grid bias or measurement artifacts. Additionally, unlike optimization-based methods, this approach does not introduce errors between relaxed functional and original functional.

We first derive the gradient flow corresponding to our model. To simplify the model derivation, we consider $\alpha(u)$ as a weighting function in the subsequent analysis, thereby yielding the Euler-Lagrange equation that corresponds to the mixed geometry information model 
\begin{equation}
	\begin{aligned}
		E^{\prime}\left(u\right)=& -\nabla \cdot\left(\alpha\left(1+|\nabla u|^2\right)^{-\frac{1}{2}} \nabla u\right) \\
		& +2 b \nabla \cdot\left(\frac{1}{\sqrt{1+|\nabla u|^2}}\left[ \left(\mathbf{I}-\mathbf{P}\right)\nabla\left(\kappa \sqrt{1+|\nabla u|^2}\right)\right]\right) \\
		& -b \nabla \cdot\left(\kappa^2\left(1+|\nabla u|^2\right)^{-\frac{1}{2}} \nabla u\right)+\eta\left(1-\frac{f}{u}\right) \\
		=& 0,
	\end{aligned}
	\label{E-L_eq}
\end{equation}
where $\kappa=\nabla \cdot \frac{\nabla \phi}{|\nabla \phi|}$, $\mathbf{I}(x)=x$, $\mathbf{P}(x)=\frac{x\cdot\nabla u}{1+|\nabla u|^2}\nabla u$. The corresponding gradient descent flow is
\begin{equation}
	\frac{\partial u}{\partial t}=-E^{\prime}\left(u\right)
	\label{evolutionEquation}
\end{equation}
with Neumann boundary condition on $\partial\Omega$
\begin{equation}
	\frac{\partial u}{\partial \mathbf{n}}=0,
\end{equation}
where $\mathbf{n}$ is the outward normal derivative at the boundary. 

From formula (\ref{E-L_eq}) we can see that $\frac{\alpha(u)+b \kappa^2}{\sqrt{1+|\nabla u|^2}}$ is the main term of diffusion. When $|\nabla u|$ is large, the diffusion coefficient is relatively small, which indicates that diffusion should be slowed down at the edges. When $|\nabla u|$ is quite small, the forward diffusion coefficient is relatively large, which indicates that rapid diffusion denoising is performed inside the image. Similar to \cite{zhu2012image}, when $|\nabla u|\ll 1$, our model will have a degenerate form
\begin{equation}
	\frac{\partial u}{\partial t}\approx\nabla\cdot\left[\big(\alpha(u)+b\left(\Delta u\right)^2\big)\nabla u\right]-2b\Delta^2u-\eta\left(1-\frac{f}{u}\right),
\end{equation}
which manifests as anisotropic diffusion and slight biharmonic diffusion. Due to the fact of $\alpha\gg b$, anisotropic diffusion plays a dominant role in the inner region.

In addition, for non-smooth boundaries, due to the existence of the denominator part $\sqrt{1+|\nabla u|^2}$, this will lead to a reduction in forward heat diffusion efficiency. However, the phenomenon of non-smooth boundaries still exists. In our model (\ref{E-L_eq}), due to the combination of the curvature term and the surface area term, it can be exactly eliminated with $\sqrt{1+|\nabla u|^2}$. This is crucial for the backward diffusion of the model. At this time, for the rough edges of the image, $\nabla\kappa$ will dominate the diffusion process of the model, causing the boundaries sharper and the interior smoother.

\section{Numerical Implementation}
\label{sec:3}
In this section, we first introduce the numerical discretization process. To overcome the time step limitations in the finite difference method,
we employ the unconditionally stable AOS and SAV algorithms.

\subsection{Operator discretization}
\label{sec:3subsec:2}
We proceed by taking time as the evolution parameter for advancing the fourth-order equation to facilitate the explicit update of $u$. Initially, we assume the image size is $M\times N$. Let $\tau$ represent the time step and define the grid as follows
\begin{equation}
	\begin{aligned}
		x_i&=i \Delta x \quad i=1, \cdots, M, \\
		y_j&=j \Delta y \quad j=1, \cdots, N, \\
		t_n&=n \tau \mspace{29mu} n=1, \cdots.
	\end{aligned}
\end{equation}
Generally $\Delta x=1$ and $\Delta y=1$. Denoting $u_{i,j}$ as the value of $u$ at $(x_i,y_j)$, various approximations of the derivative in the $x$ direction and the $y$ direction are listed
\begin{equation}
	\begin{aligned}
		& \Delta_{ \pm}^x u_{i, j}= \pm\left(u_{i \pm 1, j}-u_{i, j}\right), \\
		& \Delta_{ \pm}^y u_{i, j}= \pm\left(u_{i, j \pm 1}-u_{i, j}\right), \\
		& \Delta_c^x u_{i, j}=\frac{u_{i+1, j}-u_{i-1, j}}{2}, \\
		& \Delta_c^y u_{i, j}=\frac{u_{i, j+1}-u_{i, j-1}}{2}.
	\end{aligned}
\end{equation}
We exclusively provide the discrete scheme for the second term in (\ref{E-L_eq}), with a similar discretization approach applicable to the remaining terms. The second term is reformulated as
\begin{equation}
	V := \left(V_1,V_2\right)=\frac{1}{\sqrt{1+|\nabla u|^2}}\Big( \left(\mathbf{I}-\mathbf{P}\right)\nabla\big(\kappa \sqrt{1+|\nabla u|^2}\big)\Big),
\end{equation}
where the forms of $V_1$ and $V_2$ are
\begin{equation}
	\begin{aligned}
		& V_1=\frac{1}{\sqrt{1+|\nabla u|^2}}\left(\kappa \sqrt{1+|\nabla u|^2}\right)_x-\frac{\nabla\left(\kappa \sqrt{1+|\nabla u|^2}\right) \cdot \nabla u}{\sqrt{1+|\nabla u|^2}} u_x, \\
		& V_2=\frac{1}{\sqrt{1+|\nabla u|^2}}\left(\kappa \sqrt{1+|\nabla u|^2}\right)_y-\frac{\nabla\left(\kappa \sqrt{1+|\nabla u|^2}\right) \cdot \nabla u}{{\sqrt{1+|\nabla u|^2}}^3} u_y.
	\end{aligned}
\end{equation}
Furthermore, we list an approximation of $\nabla\cdot V$
\begin{equation}
	\nabla\cdot V= \left.\frac{\partial V_1}{\partial x}\right|_{i, j}+\left.\frac{\partial V_2}{\partial y}\right|_{i, j}=\left.V_1\right|_{i+\frac{1}{2}, j}-\left.V_1\right|_{i-\frac{1}{2}, j}+\left.V_2\right|_{i, j+\frac{1}{2}}-\left.V_2\right|_{i, j-\frac{1}{2}}.
\end{equation}
Therefore, we need to calculate $\left.V_1\right|_{i+\frac{1}{2}, j}$, $\left.V_1\right|_{i-\frac{1}{2}, j}$, $\left.V_2\right|_{i, j+\frac{1}{2}}$, $\left.V_2\right|_{i, j-\frac{1}{2}}$. Taking $\left.V_1\right|_{i+\frac{1}{2}, j}$ as an example, the discretization of others is omitted for simplicity. Let $\Psi:=\kappa \sqrt{1+|\nabla u|^2}$, and presented below is the approximate scheme for $V_1$ at $\left(i+\frac{1}{2}, j\right)$.
\begin{equation}
	\begin{aligned}
		u_x&=\Delta_{+}^xu_{i,j},\\
		u_y&=\operatorname{minmod}\left(\Delta_{c}^yu_{i,j},\Delta_{c}^yu_{i+1,j}\right),\\
		\nabla u&=\sqrt{1+\left(\Delta_{+}^x u_{i, j}\right)^2+\left(\operatorname{minmod}\left(\Delta_c^y u_{i+1, j}, \Delta_c^y u_{i, j}\right)\right)^2},\\
		\Psi_x&=\Delta_+^x\left(\kappa \sqrt{1+|\nabla u|^2}\right)_{i,j},\\
		\Psi_y&=\operatorname{minmod}\left(\Delta_{c}^y\left(\kappa \sqrt{1+|\nabla u|^2}\right)_{i,j},\Delta_{c}^y\left(\kappa \sqrt{1+|\nabla u|^2}\right)_{i+1,j}\right),\\
	\end{aligned}
\end{equation}
where the minmod function is used to increase the anisotropy of the equation
\begin{equation}
	\operatorname{minmod}(a, b)=\frac{\operatorname{sgn}(a)+\operatorname{sgn}(b)}{2} \min (|a|,|b|).
\end{equation}
The approximation of $\nabla u$ and $\kappa$ at $\left(x_i,y_j\right)$ can be expressed as
\begin{equation}
	\begin{aligned}
		\nabla u=&\sqrt{\left(\Delta_c^x u_{i, j}\right)^2+\left(\Delta_c^y u_{i, j}\right)^2},\\
		\kappa=&\Delta_{-}^x\left(\frac{\Delta_{+}^x u_{i, j}}{\sqrt{1+\left(\Delta_{+}^x u_{i, j}\right)^2+\left(\operatorname{minmod}\left(\Delta_c^y u_{i+1, j}, \Delta_c^y u_{i, j}\right)\right)^2}}\right)\\
		&+\Delta_{-}^y\left(\frac{\Delta_{+}^y u_{i, j}}{\sqrt{1+\left(\Delta_{+}^x u_{i, j}\right)^2+\left(\operatorname{minmod}\left(\Delta_c^x u_{i, j+1}, \Delta_c^x u_{i, j}\right)\right)^2}}\right).
	\end{aligned}
\end{equation}

\subsection{Numerical Discretization for Additive Operator Splitting Algorithm}
\label{sec:3subsec:3}
Let $W=\left(W_1, W_2\right):=\frac{\alpha(u)+b \kappa^2}{\sqrt{1+|\nabla u|^2}} \nabla u$ and $u_{i,j}^n$ represents the $n$th update result of $u_{i,j}$, then the evolution process can be expressed as
\begin{equation}
	\begin{aligned}
		\frac{u_{i,j}^{n+1}-u_{i,j}^{n}}{\tau}&=2b\left(\left.V_1\right|_{i+\frac{1}{2}, j}-\left.V_1\right|_{i-\frac{1}{2}, j}+\left.V_2\right|_{i, j+\frac{1}{2}}-\left.V_2\right|_{i, j-\frac{1}{2}}\right)\\
		&+\left.W_1\right|_{i+\frac{1}{2}, j}-\left.W_1\right|_{i-\frac{1}{2}, j}+\left.W_2\right|_{i, j+\frac{1}{2}}-\left.W_2\right|_{i, j-\frac{1}{2}}-\eta\left(1-\frac{f_{i,j}}{u_{i,j}^n}\right),
	\end{aligned}
\end{equation}
where $i=1\cdots M$, $j=1\cdots N$. Discretization of Neumann boundary conditions is given by
\begin{equation}
	u_{0, j}^n=u_{1, j}^n, \quad u_{M+1, j}^n=u_{M, j}^n, \quad u_{j, 0}^n=u_{j, 1}^n,\quad
	u_{i, N+1}^n=u_{i, N}^n.
\end{equation}

The explicit scheme evolves in a small time step size due to the constraints imposed by CFL conditions. This restriction diminishes computational efficiency making iteration in this scheme unfeasible.

While the theoretical foundation of model (\ref{functional}) effectively preserves the texture features, the incorporation of high-order nonlinear terms introduces substantial challenges in devising efficient numerical algorithms. Addressing the numerical instability associated with the explicit finite difference method and aiming to retain disorder in diffusion directions, we employ the additive operator splitting (AOS) algorithm as proposed in \cite{weickert1998efficient} for the numerical solution of our model.

To begin, we rewrite the evolution equation (\ref{evolutionEquation}) as
\begin{equation}
	\frac{\partial u}{\partial t}=\nabla \cdot \left(g(u)\nabla u\right) +F(u),
	\label{scheme-AOS}
\end{equation}
where 
\begin{equation}
	\begin{gathered}
		g(u):=\frac{\alpha(u)+b \kappa^2}{\sqrt{1+|\nabla u|^2}}, \\
		F(u)=-2 b \nabla \cdot\Big(
		\frac{1}{\sqrt{1+|\nabla u|^2}}\Big( \left(\mathbf{I}-\mathbf{P}\right)\nabla\big(\kappa \sqrt{1+|\nabla u|^2}\big)\Big)
		-\eta(1-\frac{f}{u})\Big).
	\end{gathered}
	\label{gF_AOS}
\end{equation}

The AOS algorithm decomposes the original diffusion into horizontal and vertical directions diffusion, then obtains $u_1^{n+1}$ and $u_2^{n+1}$ from $u^n$, and finally uses the average value of $u_1^{n+1}$ and $u_2^{n+1}$ as the iteratively updated $u^{n+1}$. From the form of (\ref{gF_AOS}), we can easily calculate the diffusion matrices $D_x^n$ and $D_y^n$ in the $x$ and $y$ directions of the size $M\times N$ image, taking the $x$ direction as an example:
\begin{equation}
	D_{x,i}^{n}=\left[\begin{array}{ccccc}
		a_1^i & b_1^i & & & \\
		c_2^i & a_2^i & b_2^i & & \\
		& \ddots & \ddots & \ddots & \\
		& & c_{N-1}^i & a_{N-1}^i & b_{N-1}^i \\
		& & & c_N^i & a_N^i
	\end{array}\right],
	\label{D_AOS}
\end{equation}
where boundary conditions are correspondingly defined as $u_{i, 0}^n=u_{i, 1}^n, u_{i, N+1}^n=u_{i, N}^n$, then one can derive each element in the matrix accordingly
\begin{equation}
	\left\{\begin{array}{l}
		a_k^i=\left\{\begin{array}{l}
			-g_{i,\frac{3}{2}}^n, \mspace{137mu} k=1, \\
			-\left(g_{i,k+\frac{1}{2}}^n+g_{i, k-\frac{1}{2}}^n\right), \quad k=2, \cdots, N-1, \\
			-g_{i,N-\frac{1}{2}}^n, \mspace{113mu} k=N,
		\end{array}\right. \\
		b_k^i=g_{i,k+\frac{1}{2}}^n, \mspace{154mu} k=1, \cdots, N-1, \\
		c_k^i=g_{i,k-\frac{1}{2}}^n, \mspace{154mu} k=2, \cdots, N.
	\end{array}\right.
\end{equation}
The values $g_{i, k+\frac{1}{2}}$ and $g_{i,k-\frac{1}{2}}$ are approximated through forward and backward differences. Subsequently, we adjust the original AOS algorithm and add an additional $F^n$ term. Upon acquiring the diffusion matrix, one-dimensional diffusion is independently applied to the rows and columns of $u^n$ for $K$ iterations.
\begin{equation}
	u^{n+1}=\frac{1}{2}\Sigma_{l}\left[I-2\tau D_l^n\left(u^n\right)\right]^{-1}\left(u^n+F^n\right),
\end{equation}
where $l\in\left\{x,y\right\}$ and $u^{n+1}$ can be solved in an efficient way.
\begin{equation}
	\begin{aligned}
		& \left(I-2 \tau D_x^n\right) u_1^{n+1}=u^n+F^n, \\
		& \left(I-2 \tau D_y^n\right) u_2^{n+1}=u^n+F^n.
	\end{aligned}
	\label{diffusionmatrix-AOS}
\end{equation}
Following the utilization of the Thomas algorithm for the alternate computation of two variables, the denoising result for iterative updates is derived by averaging the values of $u_1^{n+1}$ and $u_2^{n+1}$
\begin{equation}
	u^{n+1}=\frac{1}{2}\left(u_1^{n+1}+u_2^{n+1}\right).
\end{equation}

After the tridiagonal matrices $D_x^n$ and $D_y^n$ are obtained through (\ref{D_AOS}), the problem of image denoising is transformed into the problem of solving linear system. The AOS algorithm possesses absolute stability, consistently yielding outstanding experimental results even under large time step sizes. However, in order to maintain the accuracy of the algorithm, it is common to set a smaller time step size. Consequently, in experiments, a delicate balance must be struck between solution accuracy and computational efficiency. The choice of an appropriate time step size involves a tradeoff. Algorithm \ref{algorithm_AOS} outlines the AOS algorithm for solving (\ref{scheme-AOS}).
\begin{algorithm}
	\caption{The AOS for mixed geometry information model}
	\label{algorithm_AOS}
	\begin{algorithmic}[1]
		\Require
		Noisy image $f$, iterations $K$, time step size $\tau$.
		\Ensure Denoising result $u^K$ of step $K$.
		\State Initialization.	$u^0 \longleftarrow f$\;
		\For{$n = 0\cdots K$}
		\For{$j = 1\cdots N$}
		\State Compute $a_k^{j}$, $b_k^{j}$, $c_k^{j}$, get the diffusion matrix $D_{y}^{n}$\;
		\State Solving $\left(I-2 \tau D_{y}^n\right) u_{2, j}^{n+1}=u_j^n+F_j^n$ using Thomas algorithm\;
		\EndFor
		\For{$i= 1\cdots M$}
		\State	Compute $a_k^{i}$, $b_k^{i}$, $c_k^{i}$, get the diffusion matrix$D_{x}^{n}$\;
		\State Solving  $\left(I-2 \tau D_{x, i}^n\right) \left(u_{1, i}^{n+1}\right)^T=\left(u_i^n\right)^T+\left(F_i^n\right)^T$ using Thomas algorithm\;
		\EndFor
		\State $u_{i,j}^{n+1}=\frac{1}{2}\left(u_1^{n+1}+u_2^{n+1}\right)$\;
		\EndFor
		\State	Calculate relevant quality evaluation indicators and count calculation time.
	\end{algorithmic}
\end{algorithm}
\begin{theorem}For any time step $\tau \textgreater 0$, the scheme (\ref{diffusionmatrix-AOS}) in AOS algorithm is absolutely stable.
\end{theorem}
\begin{pproof}
	From the form of $g$ in (\ref{gF_AOS}), we know $g_{i,j}\geq 0$, for $i=1,\cdots M$, $j=1,\cdots N$. Then the main diagonal elements in the matrix $D^n_{x,i}$ corresponding to (\ref{D_AOS}) are negative values, and the subdiagonal elements are positive values. And we can easily see that $D^n_{x,i}$ is a symmetric matrix. In addition, we need to note that $D^n_{x,i}$ is not only a weak diagonally dominant matrix, but also a irreducibly matrix. Applying the Cottle theorem \cite{MR3396730}, $D^n_{x,i}$ has negative eigenvalues $\lambda \textless 0$.
	
	Let $Q^n$ represents the matrix $(I-2 \tau D_x^{n})^{-1}$. The eigenvalue of matrix  is  $Q^n \in(0,1)$. On the other hand, we know the fact that the sum of eigenvalues is equal to the sum of the diagonal elements. So the spectral radius of $D^n_{x,i}$ is strictly greater than 0 due to the fact that the main diagonal elements of $D^n_{x,i}$ are strictly less than 0. Then denoting $L\in\left(0,1\right)$ as the spectral radius of $Q^n$. For any $n,p$
	\begin{equation}
		\begin{aligned}
			\left\|u_1^{n+p}-u_1^{n}\right\|=&\left\|\left(Q^{n+p-1}\cdots Q^n-I\right)\left(Q^{n-1}\cdots Q^0\right)\left(u^0+F^0\right)\right\|\\
			\leq&L^{n}\left\|\left(u^0+F^0\right)\right\|.
		\end{aligned}
	\end{equation}
	This inequality indicates $u_1^n$ is a convergent sequence. Similarly, the convergence of $u_2^n$ can be obtained.
	Thus the semi-implicit AOS algorithm \ref{algorithm_AOS} corresponding to (\ref{functional}) is unconditionally stable for any $\tau \textgreater 0$.
\end{pproof}
\begin{remark}
	\normalfont
	While there is an additional term involving $\nabla u$ in the $\mathbf{P}$ operator in (\ref{gF_AOS}), this term cannot be incorporated into the operator $g$. On the one hand, the $\mathbf{P}$ operator fails to ensure the positive definiteness of the coefficient of $\left(\nabla u\right)$, leading to the inability to satisfy the prerequisite conditions for $g$ in the $\mathbf{P}$ operator. Hence unconditional stability cannot be guaranteed, resulting in oscillatory behavior during the iterative process. On the other hand, overly complex diffusion coefficients may, to some extent, conflict with each other, introducing ambiguity in the diffusion process. This underscores that the AOS framework fundamentally operates as a second-order solution framework. Consequently, the inclusion of high-order derivatives in the diffusion coefficients for both the $x$ and $y$ directions should be approached cautiously. An excessive presence of high-order terms may potentially lead to algorithmic malfunction.
\end{remark}

\subsection{Numeric Discretization of Scalar Auxiliary Variable Scheme}
\label{sec:3subsec:4}
While the AOS algorithm has been used to successfully solve numerous nonlinear models, it is accompanied by the drawbacks of reduced accuracy and `streaking artifacts' \cite{min2014fast,alcantarilla2011fast}. It is worth noting that in order to solve the problem of energy functional minimization, Shen et al. \cite{shen2018scalar,shen2019new} transformed the gradient flow solution of the original problem by introducing a scalar auxiliary variable, and proposed the scalar auxiliary variable (SAV) method. As a burgeoning numerical algorithm in the field of image processing, SAV represents a general and efficient solution framework with simple numerical scheme, accurate calculation accuracy, and unconditional stability \cite{yao2020total,wang2022efficient,bai2023ginzburg}.

Redefine the new energy functional (\ref{functional}), which can be rewritten as:

\begin{equation}
	\varepsilon\left[u\right] = \frac{\gamma}{2}\left(u,Lu\right)+\varepsilon_1\left[u\right],
	\label{functional_SAV}
\end{equation}
where $L$ is a non-negative symmetric linear operator, typically independent of the function $u$ and $\varepsilon_1\left[u\right]$ containing a nonlinear term related to $u$. As all terms in problem (\ref{functional}) are nonlinear operators, we take $\varepsilon_1\left[u\right]$ equal to the total energy functional subtract linear term inner product. This choice ensures that $\varepsilon\left[u\right]$ is still equal to the total energy functional, allowing them to be harmonized into a unified form. Furthermore, the $L^2$ gradient flow corresponding to the minimization of the functional (\ref{functional_SAV}) is
\begin{equation}
	\frac{\partial u}{\partial t}=-\frac{\delta \varepsilon}{\delta u}.
\end{equation}
By the chain rule, we get that the derivative of energy with respect to time 
\begin{equation}
	\frac{\mathrm{d} \varepsilon[u]}{\mathrm{d} t}=\frac{\delta \varepsilon}{\delta u} \cdot \frac{\partial u}{\partial t}=\left(\frac{\delta \varepsilon}{\delta u},-\frac{\delta \varepsilon}{\delta u}\right) = -\left\|\frac{\delta \varepsilon}{\delta u}\right\|_2^2 \leq 0.
\end{equation}
Therefore, the gradient descent method effectively maintains the reduction of energy during the evolution process. Obviously, the objective functional possesses a well-defined lower bound, so there exists a positive constant $C \textgreater 0$  such that $\varepsilon_1\left[u\right]=E\left(u\right)-\frac{\gamma}{2}\left(u,Lu\right)+C\textgreater 0$. To facilitate analysis, we introduce the scalar auxiliary variable $r=\sqrt{\varepsilon_1}$ and present the equivalent form of the gradient flow within the SAV framework
\begin{subequations}
	\begin{empheq}[left=\hspace{-2.55cm}\empheqlbrace]{alignat=3}
		\frac{\partial u}{\partial t} & =-\mu, \label{con_SAV_1}\\
		\mu & =\gamma L u+\frac{r}{\sqrt{\varepsilon_1[u]}} \varepsilon_1^{\prime}\left(u\right), \label{con_SAV_2}\\
		r_t & =\frac{1}{2 \sqrt{\varepsilon_1[u]}} \int_{\Omega} \varepsilon_1^{\prime}\left(u\right) u_t \mathrm{~d} x.\label{con_SAV_3}
	\end{empheq}
	\label{con_SAV}%
\end{subequations}
Next, we introduce the first-order scheme of SAV
\begin{subequations}
	\begin{empheq}[left=\empheqlbrace]{alignat=3}
		&\frac{u^{n+1}-u^n}{\tau}=-\mu^{n+1}, \label{firstorder_SAV_1}\\
		&\mu^{n+1}=\gamma L u^{n+1}+\frac{r^{n+1}}{\sqrt{\varepsilon_1}} \varepsilon_1^{\prime}\left(u^n\right), \label{firstorder_SAV_2}\\
		&r^{n+1}-r^n=\frac{1}{2\sqrt{\varepsilon_1}} \int_{\Omega} \varepsilon_1^{\prime}\left(u^n\right)\left(u^{n+1}-u^n\right) 			\mathrm{d} x. \label{firstorder_SAV_3}
	\end{empheq}
	\label{firstorder_SAV}%
\end{subequations}

Although the update of $u$ in (\ref{firstorder_SAV}) may appear intricate, the computational cost primarily focuses on solving a linear system with constant coefficients, and the associated calculation cost is comparatively modest. Moreover, we can efficiently address (\ref{firstorder_SAV}) through the following process:

Substitute (\ref{firstorder_SAV_2}) and (\ref{firstorder_SAV_3}) into (\ref{firstorder_SAV_1}), leading to

\begin{equation}
	\left(I+\tau \gamma L\right)u^{n+1} + \frac{\tau}{2}b^n\left(b^n,u^{n+1}\right)=c^n,
	\label{un+1_SAV}
\end{equation}
where 
\begin{equation}
	\begin{aligned}
		I\left(x\right)&=x,\\
		b^n&=\frac{\varepsilon_1^{\prime}\left(u^n\right)}{\sqrt{\varepsilon_1}},\\
		c^n&=u^n-\tau r^nb^n+\frac{\tau}{2}b^n\left(b^n,u^n\right).
	\end{aligned}
	\label{bncn}
\end{equation}
Multipling the (\ref{un+1_SAV}) by the $b^n$ and integrating, we obtain
\begin{equation}
	\left(I+\tau \gamma L\right)\left(b^n,u^{n+1}\right) + \frac{\tau}{2}\left\|b^n\right\|_2^2\left(b^n,u^{n+1}\right)=\left(b^n,c^n\right).
\end{equation}
Then there holds
\begin{equation}
	\left(b^n,u^{n+1}\right)=\frac{\left(b^n,\left(I+\tau \gamma L\right)^{-1}c^n\right)}{1+\frac{\tau}{2}\left(b^n,\left(I+\tau \gamma L\right)^{-1}b^n\right)}.
	\label{unbn+1_SAV}
\end{equation}
Finally, we substitute (\ref{unbn+1_SAV}) back into (\ref{un+1_SAV}) to get $u^{n+1}$, which gives
\begin{equation}
	u^{n+1}=\left(I+\tau \gamma L\right)^{-1}\big(c^n-\frac{\tau}{2}b^n(b^n,u^{n+1})\big).
\end{equation}
In summary, the first-order scheme of the SAV algorithm can be summarized as Algorithm \ref{algorithm_SAV}.
\begin{algorithm}
	\caption{The SAV with an adaptive time stepping strategy for mixed geometry information model}
	\label{algorithm_SAV}
	\begin{algorithmic}[1]
		\Require
		Noisy image $f$, time step size $\tau$.
		\Ensure Final denoised result $u^{F}$.
		\State Initialization.	$u^0 \longleftarrow f$\;
		\For{$n = 0\cdots $}
		\State Compute $b^n$ and $c^n$ through the formula (\ref{bncn}) or (\ref{bncn_secondorder})
		\State Compute $\left(b^n,u^{n+1}\right)$ through the formula (\ref{unbn+1_SAV}) or (\ref{unbn+1_secondorder})
		\State Compute $u^{n+1}$ through the formula (\ref{un+1_SAV}) or (\ref{un+1_secondorder})
		\State Compute $\tau_{n+1}$ through th formula (\ref{Adaptivetimestep})
		\State End until some stopping criterion is reached, get the denoising result $u^{F}$
		\EndFor
		\State	Calculate relevant quality evaluation indicators and count calculation time.
	\end{algorithmic}
\end{algorithm}

SAV approaches of second-order or higher-order schemes can also be easily constructed, typically exhibiting energy stability. Therefore the selection of the time step size becomes less critical, as it does not necessitate adherence to the CFL condition. In what follows, we introduce a second-order scheme within the SAV framework, employing the Crank-Nicolson scheme.

\begin{subequations}
	\begin{empheq}[left=\empheqlbrace]{alignat=3}
		&\frac{u^{n+1}-u^n}{\tau}=-\mu^{n+\frac{1}{2}}, \label{secondorder_SAV_1}\\
		&\mu^{n+\frac{1}{2}}=\gamma L \left[\frac{1}{2}\left(u^n+u^{n+1}\right)\right]+\frac{r^{n+1}+r^n}{2\sqrt{\varepsilon_1\left[\widetilde{u}^{n+\frac{1}{2}}\right]}} \varepsilon_1^{\prime}\left(\widetilde{u}^{n+\frac{1}{2}}\right), \label{secondorder_SAV_2}\\
		&r^{n+1}-r^n=\int_{\Omega}\frac{\varepsilon_1^{\prime}\left(\widetilde{u}^{n+\frac{1}{2}}\right)}{2\sqrt{\varepsilon_1\left[\widetilde{u}^{n+\frac{1}{2}}\right]}}  \left(u^{n+1}-u^n\right) \mathrm{~d} x,\label{secondorder_SAV_3}		
	\end{empheq}
	\label{secondorder_SAV}%
\end{subequations}%
where the value of $\widetilde{u}^{n+\frac{1}{2}}$ can be approximated using an explicit difference method, as exemplified by the following formulation
\begin{equation}
	\widetilde{u}^{n+\frac{1}{2}}=\frac{1}{2}\left(3 u^n-u^{n-1}\right).
\end{equation}
Then the iterative update of $u^n$ can be derived in a similar process in the first-order scheme. First, substitute (\ref{secondorder_SAV_2}) and (\ref{secondorder_SAV_3}) into (\ref{secondorder_SAV_1}), we get
\begin{equation}
	\left(I+\frac{\tau}{2}\gamma L\right)u^{n+1}+\frac{\tau}{4}b^n\left(b^n,u^{n+1}\right)=c^n,
	\label{un+1_secondorder}
\end{equation}
where
\begin{equation}
	\begin{aligned}
		b^n&=\frac{\varepsilon_1^{\prime}\left(\widetilde{u}^{n+\frac{1}{2}}\right)}{2\sqrt{\varepsilon_1\left[\widetilde{u}^{n+\frac{1}{2}}\right]}},\\
		c^n&=u^n-\frac{\tau}{2}\gamma Lu^n-\tau b^nr^n+\frac{\tau}{4}b^n\left(b^n,u^n\right).
	\end{aligned}
	\label{bncn_secondorder}
\end{equation}
Similarly, multipling (\ref{bncn_secondorder}) by $b^n$ and integrating, then there holds
\begin{equation}
	\left(I+\frac{\tau}{2} \gamma L\right)\left(b^n,u^{n+1}\right) + \frac{\tau}{4}\left\|b^n\right\|_2^2\left(b^n,u^{n+1}\right)=\left(b^n,c^n\right).
\end{equation}
Furthermore, it arrives at
\begin{equation}
	\left(b^n,u^{n+1}\right)=\frac{\left(b^n,\left(I+\frac{\tau}{2}\gamma L\right)^{-1}c^n\right)}{1+\frac{\tau}{4}\left(b^n,\left(I+\tau \gamma L\right)^{-1}b^n\right)}.
	\label{unbn+1_secondorder}
\end{equation}
Finally, combining (\ref{un+1_secondorder}) and (\ref{unbn+1_secondorder}), it leads to the iterative update result of $u^{n+1}$
\begin{equation}
	u^{n+1}=\left(I+\frac{\tau}{2} \gamma L\right)^{-1}\left(c^n-\frac{\tau}{4}b^n\left(b^n,u^{n+1}\right)\right).
\end{equation}

During the practical evolution process, the energy functional may experience rapid decay within a short time period, followed by slight changes in the remaining periods. Therefore, it becomes imperative to opt for a small step size when the energy undergoes rapid changes and a relatively larger step size during periods of energy stability. To address this, we implement an adaptive time step strategy according to \cite{shen2019new},
\begin{equation}
	\tau^{n+1}={\rm max}\left\{\tau_{min},{\rm min}\left\{\rho\left(\frac{t o l}{e}\right)^{1 / 2} \tau^{n},\tau_{max}\right\}\right\},
	\label{Adaptivetimestep}
\end{equation}
where $\rho$ is a default safety coefficient, $tol$ is a constant representing reference tolerance and $e$ is the relative error in the iteration, $\tau_{min}$ is the minimum time step size, $\tau_{max}$ is the maximum time step size. 

We now give the unconditional energy stability properties of the numerical scheme.
\begin{theorem} The energy functional $E(u)$ in the scheme (\ref{con_SAV}) is steadily declining
	\begin{equation*}
		\frac{\mathrm{d} E(u)}{\mathrm{d} t}\leq 0.
	\end{equation*} 
\end{theorem}
\begin{pproof}
	Firstly, we multiply $\frac{\partial u}{\partial t}$ in the both side of (\ref{con_SAV_1}) and (\ref{con_SAV_2}) and intergrate on the image domain $\Omega$,
	\begin{equation}
		\left(\frac{\partial u}{\partial t},\frac{\partial u}{\partial t}\right)=-\left(\mu, \frac{\partial u}{\partial t}\right),
		\label{Thm2_1}
	\end{equation}
	\begin{equation}
		\left(\mu, \frac{\partial u}{\partial t}\right)=\gamma\left(Lu,\frac{\partial u}{\partial t}\right)+\frac{r}{\sqrt{\varepsilon_1[u]}} \int_\Omega\varepsilon_1^{\prime}\left(u\right)\frac{\partial u}{\partial t}\mathrm{~d} x.
		\label{Thm2_2}
	\end{equation}
	After multplying the $2r$ on the both side of (\ref{con_SAV_3}) and swap both sides of (\ref{Thm2_1}), we obtain
	\begin{equation}
		2rr_t+\gamma\left(Lu,\frac{\partial u}{\partial t}\right)+\left(\frac{\partial u}{\partial t},\frac{\partial u}{\partial t}\right)=0.
	\end{equation}
	By the positive definiteness of the inner product, we have
	\begin{equation}
		\frac{\partial\left(\frac{\gamma}{2}\left(Lu,u\right)+r^2-C\right)}{\partial t}\leq 0.
	\end{equation}
	Then we can easily know the fact that $\frac{\mathrm{d} E(u)}{\mathrm{d} t}\leq 0$.
\end{pproof}

Multiply the three equations of (\ref{firstorder_SAV}) by $\mu^{n+1}$, $\frac{u^{n+1}-u^n}{\tau}$, and $\frac{2r^{n+1}}{\tau}$, respectively. Then integrate (\ref{firstorder_SAV_1}) and (\ref{firstorder_SAV_2}), and sum up them. We readily derive the following theorem on energy dissipation. The proof follows from the proof given in \cite{shen2019new}.
\begin{theorem} The scheme (\ref{firstorder_SAV}) is first-order accurate and unconditionally energy stable in the sense that
	\begin{equation*}
		\begin{aligned}
			\frac{1}{\tau}&\left(\varepsilon(u^{n+1})-\varepsilon(u^{n})\right)\\
			&+\frac{1}{2}\Big(\frac{\gamma}{\tau}\left(u^{n+1}-u^n, L\left(u^{n+1}-u^n\right)\right)+\left(r^{n+1}-r^n\right)^2\Big)=-\left(\mu^{n+1},\mu^{n+1}\right)\le 0.
		\end{aligned}
	\end{equation*}
\end{theorem}

Similarly, multiply the three equations of scheme (\ref{secondorder_SAV}) by $\mu^{n+\frac{1}{2}}$, $\frac{u^{n+1}-u^n}{\tau}$, and $\frac{r^{n+1}+r^n}{\tau}$, respectively, integrate the first two equations, and finally sum up all the three equations. Using the linear and symmetric properties of $L$, we get the following theorem.
\begin{theorem} The scheme (\ref{secondorder_SAV}) is second-order accurate and unconditionally energy stable in the sense that
	\begin{equation*}
		\frac{1}{\tau}\left(\varepsilon(u^{n+1})-\varepsilon(u^{n})\right)=-\left(\mu^{n+\frac{1}{2}},\mu^{n+\frac{1}{2}}\right)\le 0.
	\end{equation*}
\end{theorem}

\section{Numerical experiment}
\label{sec:4}
In this section, we evaluate the performance of the proposed model and algorithm through numerical experiments on the multiplicative denoising problem. Specifically, we use Peak Signal to Noise Ratio(PSNR) and Structural SIMilarity(SSIM)\cite{wang2004image} as quantitative evaluation metrics to measure the quality of recovered images. PSNR and SSIM are defined as
\begin{equation}
	\begin{aligned}
		\mathrm{PSNR}&=10 \log _{10}\left(\frac{255^2}{\frac{1}{M N} \sum_{i=1}^M \sum_{j=1}^N[\hat{u}(i, j)-u(i, j)]^2}\right),\\
		\mathrm{SSIM}&=\frac{\left(2 \mu_u \mu_{\hat{u}}+c_1\right)\left(2 \sigma_{u \hat{u}}+c_2\right)}{\left(\mu_u^2 \mu_{\hat{u}}^2+c_1\right)\left(\sigma_u^2+\sigma_{\hat{u}}^2+c_2\right)},
	\end{aligned}
\end{equation}
where $u$ is the original image of size $M\times N$, $\hat{u}$ is the restored image, $\mu_u$ and $\mu_{\hat{u}}$ are mean values of $u$ and $\hat{u}$ respectively, $\sigma_u^2$ and $\sigma_{\hat{u}}^2$ represent the variances of $u$ and $\hat{u}$ respectively, $\sigma_{u \hat{u}}$ is the covariance of $u$ and $\hat{u}$, $c_1\textgreater 0$ and $c_2\textgreater 0$ are constants.

\subsection{Parameters Discussing in AOS Algorithm}
\label{sec:4subsec:1}

As mentioned in Section \ref{sec:3subsec:3}, the AOS algorithm is absolutely stable, so the time step can be selected arbitrarily in terms of stability. However, although an excessively large time step can greatly improve the computational efficiency, it also lead to too fast to diffuse and miss the best denoising result. Therefore, it is very important to find a compromise step to balance efficiency and accuracy. In order to obtain a suitable time step size, we tested the time step size $\tau$ at 4 different scales, while keeping other parameters fixed.

\begin{figure}[h]
	\centering
	\begin{minipage}{\textwidth}
		\centering
		\subfigure[$\tau=1$]{
			\includegraphics[width=0.23\textwidth]{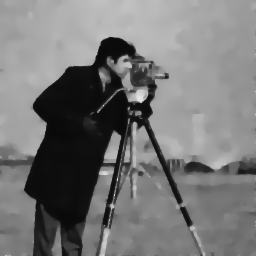}
		}
		\subfigure[$\tau=2$]{
			\includegraphics[width=0.23\textwidth]{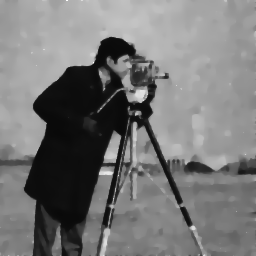}
		}
		\subfigure[$\tau=5$]{
			\includegraphics[width=0.23\textwidth]{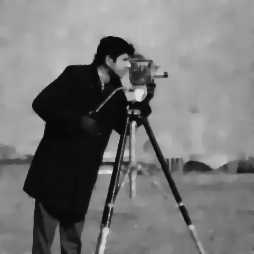}
		}
		\subfigure[$\tau=10$]{
			\includegraphics[width=0.23\textwidth]{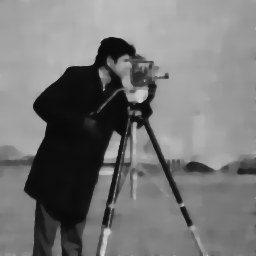}
		}
	\end{minipage}
	\caption{The denoising effect of the image with noise level $L=10$ under different time step sizes. Meanwhile, $b=0.01$, $\eta=0.01$, correspondingly. The optimal number of iterations, PSNR and SSIM values are as follows. a 76, 25.96, 0.7881, b 42, 26.02, 0.7878, c 20, 26.02, 0.7879, d 11, 25.76, 0.7749}
	\label{ParDiscussion_AOS}
\end{figure}
As observed, Fig. \ref{ParDiscussion_AOS}c and Fig. \ref{ParDiscussion_AOS}d are blurred, losing some details. Compared with $\tau=1$, $\tau=2$ can reduce the calculation time by half, while hardly losing recovery information. Therefore, we specify the time step $\tau=2$ when the noise level $L=10$.

\subsection{Parameters Discussing in SAV Algorithm}
\label{sec:4subsec:2}
In this section, we take the first-order SAV algorithm as an example to show the results. The second-order SAV algorithm obtains similar experimental results. We test an important parameter in the algorithm, the constant $C$. As mentioned above, the energy functional itself has a lower bound, which needs to be added to a constant to ensure its non-negativity. The optimization process of modified functional is equivalent to the original functional, but in numerical calculation, it may be different when $C$ changes.

\begin{figure}[h]
	\centering
	\begin{minipage}{\textwidth}
		\centering
		\subfigure[$C=10^7$]{
			\includegraphics[width=0.23\textwidth]{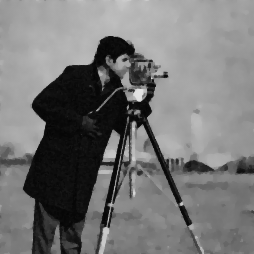}
		}
		\subfigure[$C=10^8$]{
			\includegraphics[width=0.23\textwidth]{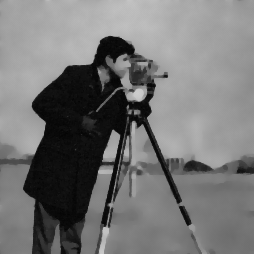}
		}
		\subfigure[$C=10^9$]{
			\includegraphics[width=0.23\textwidth]{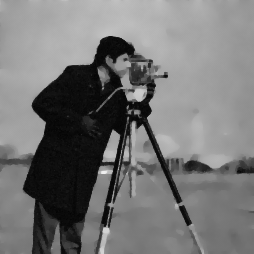}
		}
		\subfigure[$C=10^{10}$]{
			\includegraphics[width=0.23\textwidth]{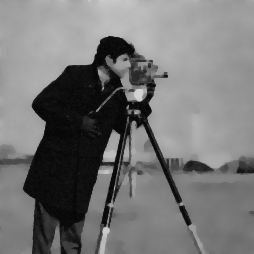}
		}
	\end{minipage}
	\caption{The denoising effect of the image with noise level $L=10$ under different time step sizes. Meanwhile, $b=0.001$, $\eta=0.15$, $\tau_{min}=0.8$, $\tau_{max}=1.2$. The CPU time, PSNR and SSIM values are as follows. a 15.9242, 26.48, 0.7948, b 18.1214, 26.39, 0.7935, c 12.1957, 26.46, 0.7733, d 11.9296, 26.31, 0.7894}
	\label{ParDiscussion_SAV_came}
\end{figure}

\begin{figure}[h]
	\centering
	\begin{minipage}{\textwidth}
		\centering
		\subfigure[$C=10^{12}$]{
			\includegraphics[width=0.23\textwidth]{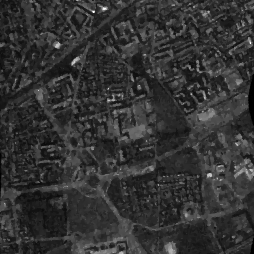}
		}
		\subfigure[$C=10^{13}$]{
			\includegraphics[width=0.23\textwidth]{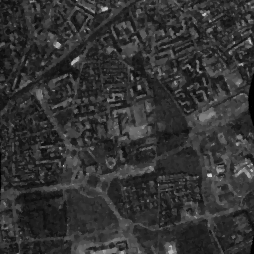}
		}
		\subfigure[$C=10^{14}$]{
			\includegraphics[width=0.23\textwidth]{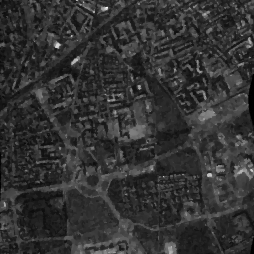}
		}
		\subfigure[$C=10^{15}$]{
			\includegraphics[width=0.23\textwidth]{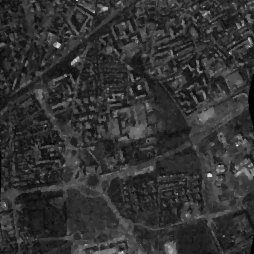}
		}
	\end{minipage}
	\caption{The denoising effect of the image with noise level $L=10$ under constant $C$. And we set $b=0.01$, $\eta=0.3$, $\tau_{min}=0.1$, $\tau_{max}=0.5$, correspondingly. The CPU time, PSNR and SSIM values are as follows. a 19.0688, 27.14, 0.8008, b 15.8645, 27.10, 0.7995, c 17.0544, 27.13 0.7999, d 18.1358, 27.10, 0.7970}
	\label{ParDiscussion_SAV_aerial}
\end{figure}
As observed, PSNR does not have much correlation with the value of $C$, so we choose $C=10^7$. But one issue that needs to be pointed out is that the denoising effect of Fig. \ref{ParDiscussion_SAV_came}d is extremely unstable. This is due to the rapid decay of the energy to a small magnitude during the evolution. Under the condition of the relative error stopping criterion, the relative error is easily less than the agreed threshold when the constant is too large, so the loop is jumped out, resulting in unsatisfactory results. In addition, unlike the arbitrary choice of $C$ in the literature \cite{wang2022efficient}, our energy functional itself has a lower bound but may have negative values. Therefore, when $C$ is too small, it will bring the singularity of the equation, and the choice of $C$ is related to the image itself. For example, when $C=10^7$, it is successful for $Cameraman$, but it will fail for $Aerial$. Therefore, combined with the consideration of non-negative auxiliary variables and the convergence of the model, we change the stopping criterion in $Aerial$ to absolute error, which can effectively resolve their contradictions.

\subsection{Denoising Performance}
\label{sec:4subsec:4}
In this subsection, we compare the proposed model with many efficient multiplicative denoising models, including AA model\cite{aubert2008variational}, DD model\cite{zhou2014doubly}, MuLog+BM3D model\cite{deledalle2017mulog}, NTV model \cite{zhao2014new}, AAFD model\cite{yao2019multiplicative}, EE model\cite{zhang2022image} which is solved by alternating direction method of multipliers. 

\begin{figure}[h]
	\centering
	\begin{minipage}{\textwidth}
		\centering
		\subfigure[Clean]{
			\includegraphics[width=0.19\textwidth]{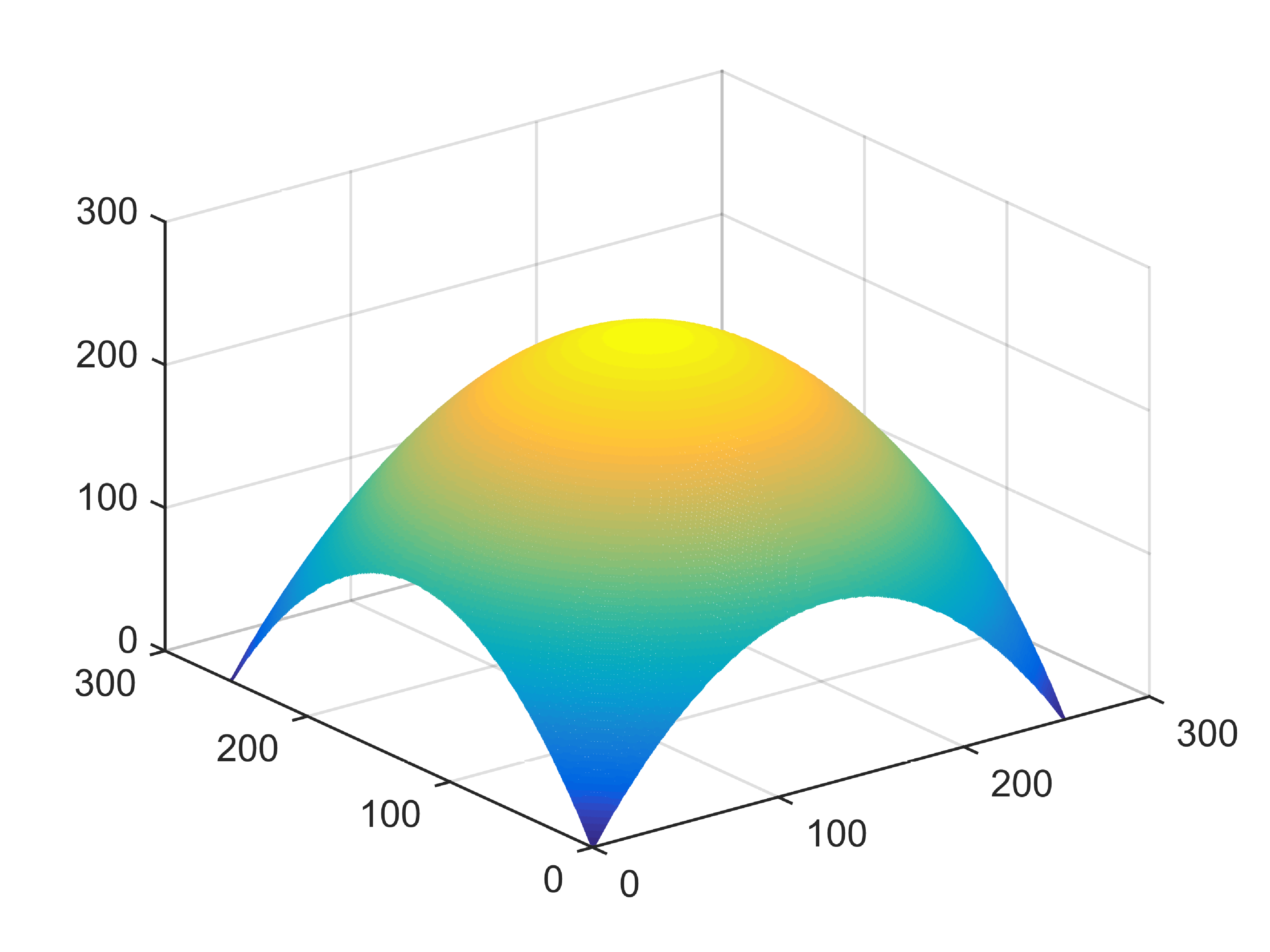}
		}
		\subfigure[Noisy]{
			\includegraphics[width=0.19\textwidth]{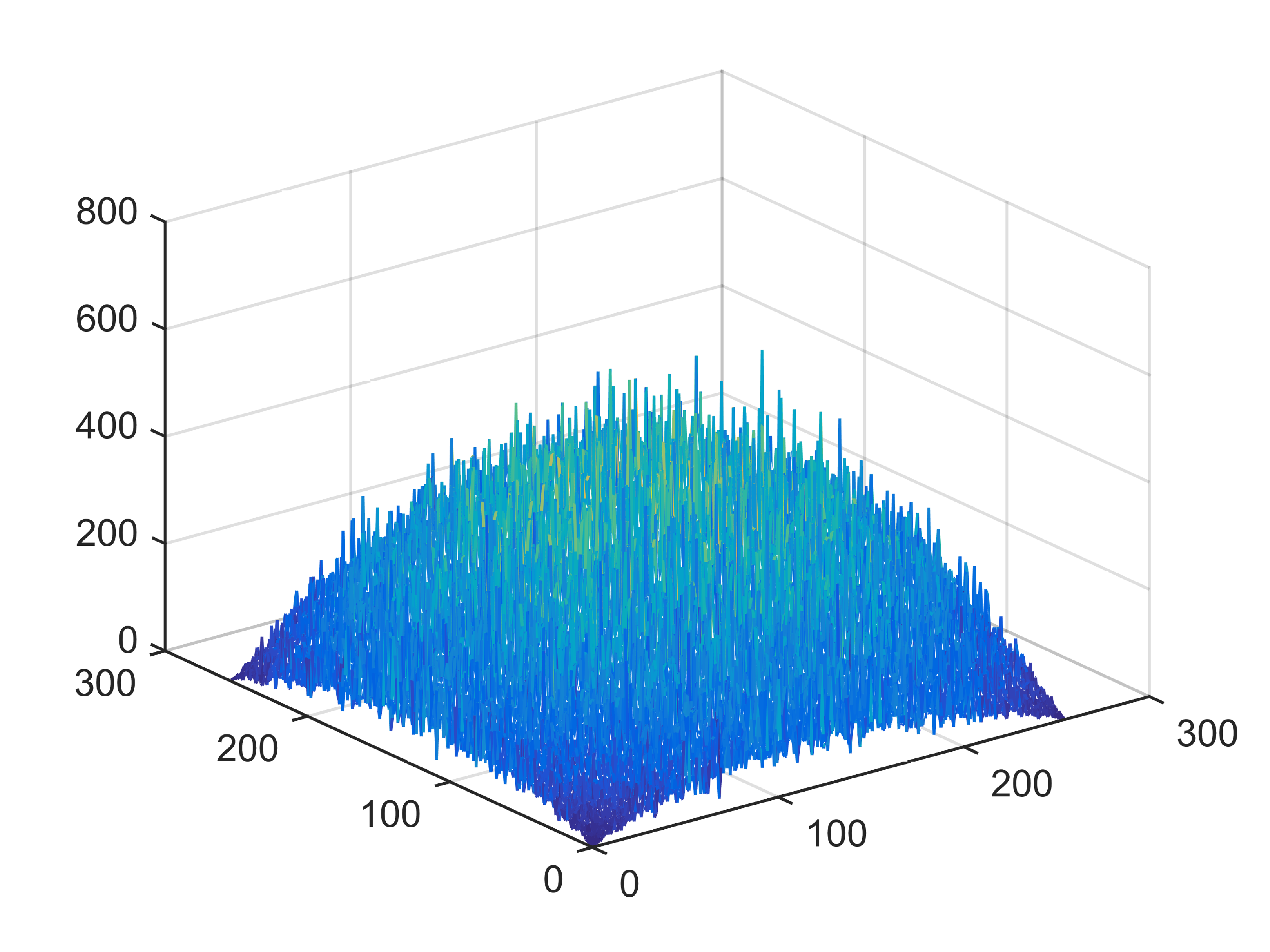}
		}
		\subfigure[AA]{
			\includegraphics[width=0.19\textwidth]{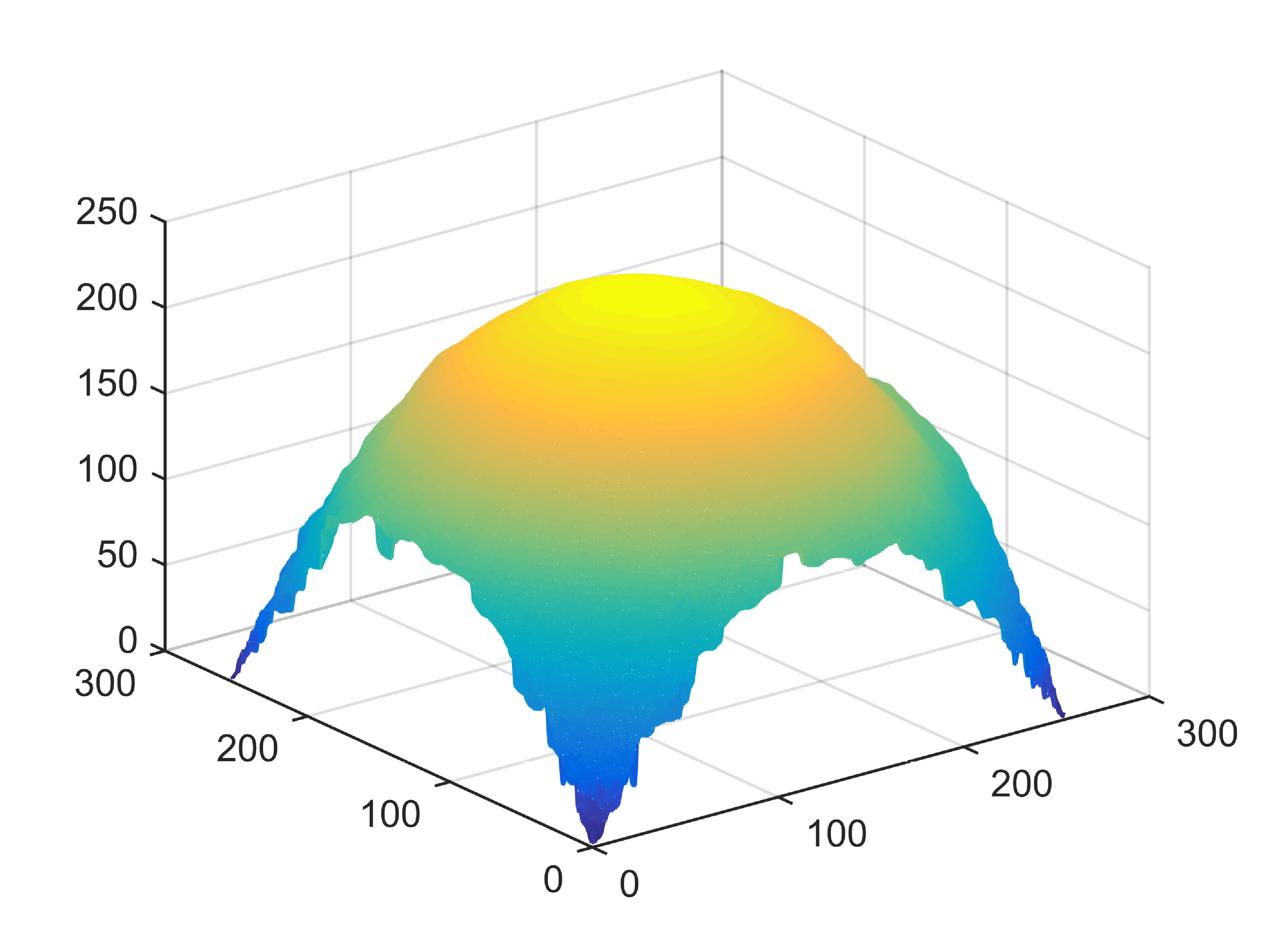}
		}
		\subfigure[DD]{
			\includegraphics[width=0.19\textwidth]{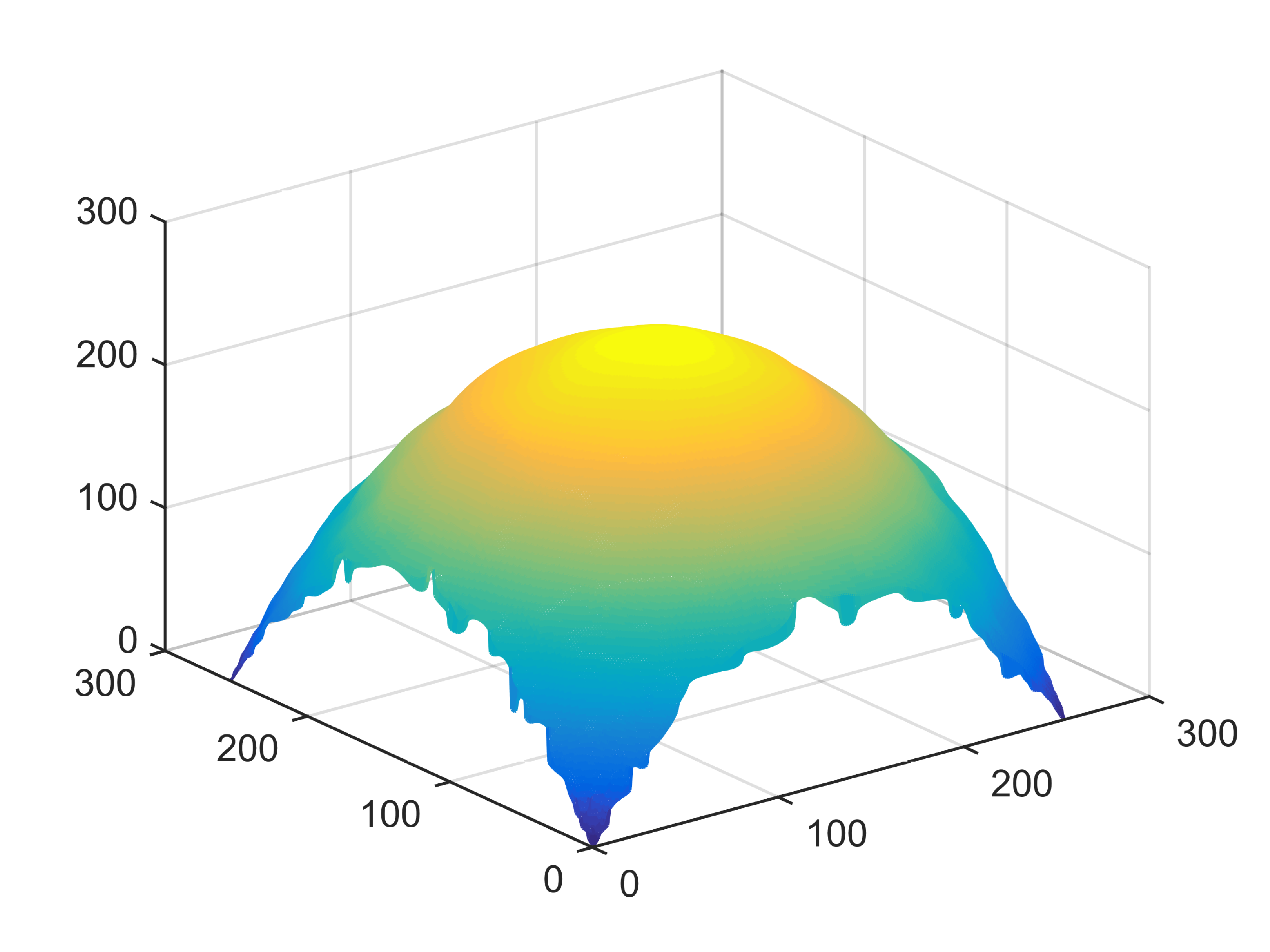}
		}
		\subfigure[MuLoG]{
			\includegraphics[width=0.19\textwidth]{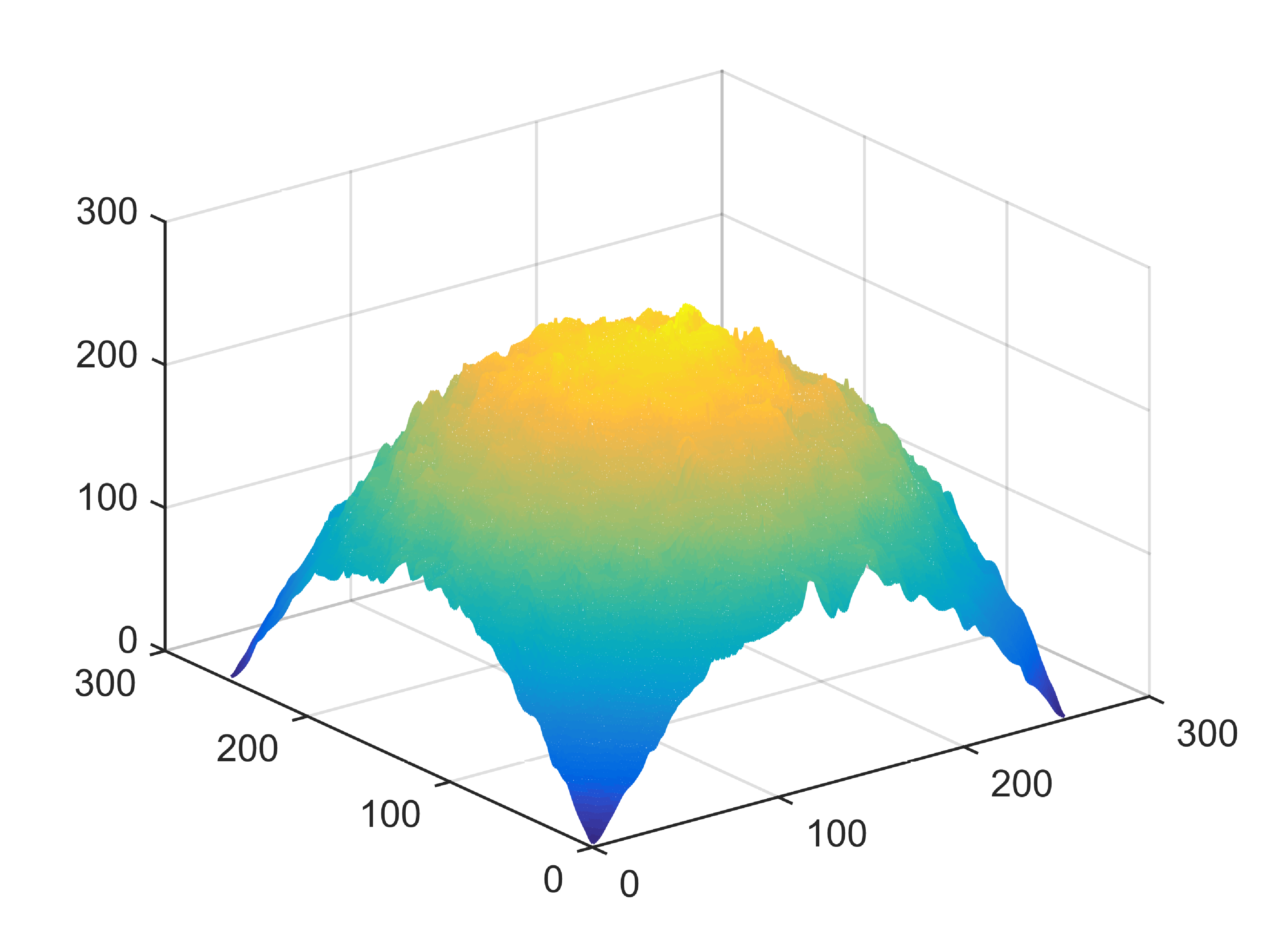}
		}
	\end{minipage}
	\begin{minipage}{\textwidth}
		\centering
		\subfigure[NTV]{
			\includegraphics[width=0.19\textwidth]{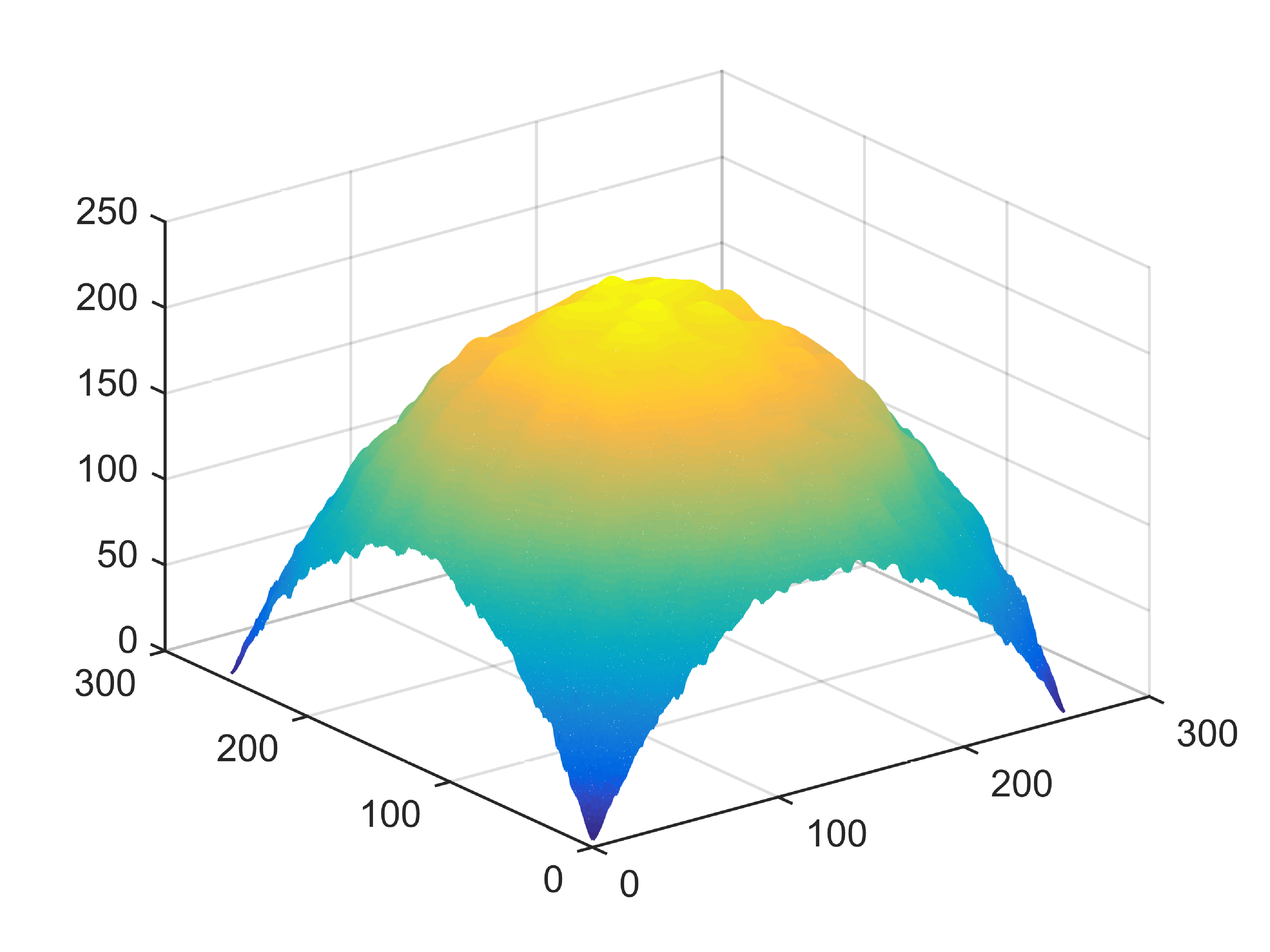}
		}
		\subfigure[EE]{
			\includegraphics[width=0.19\textwidth]{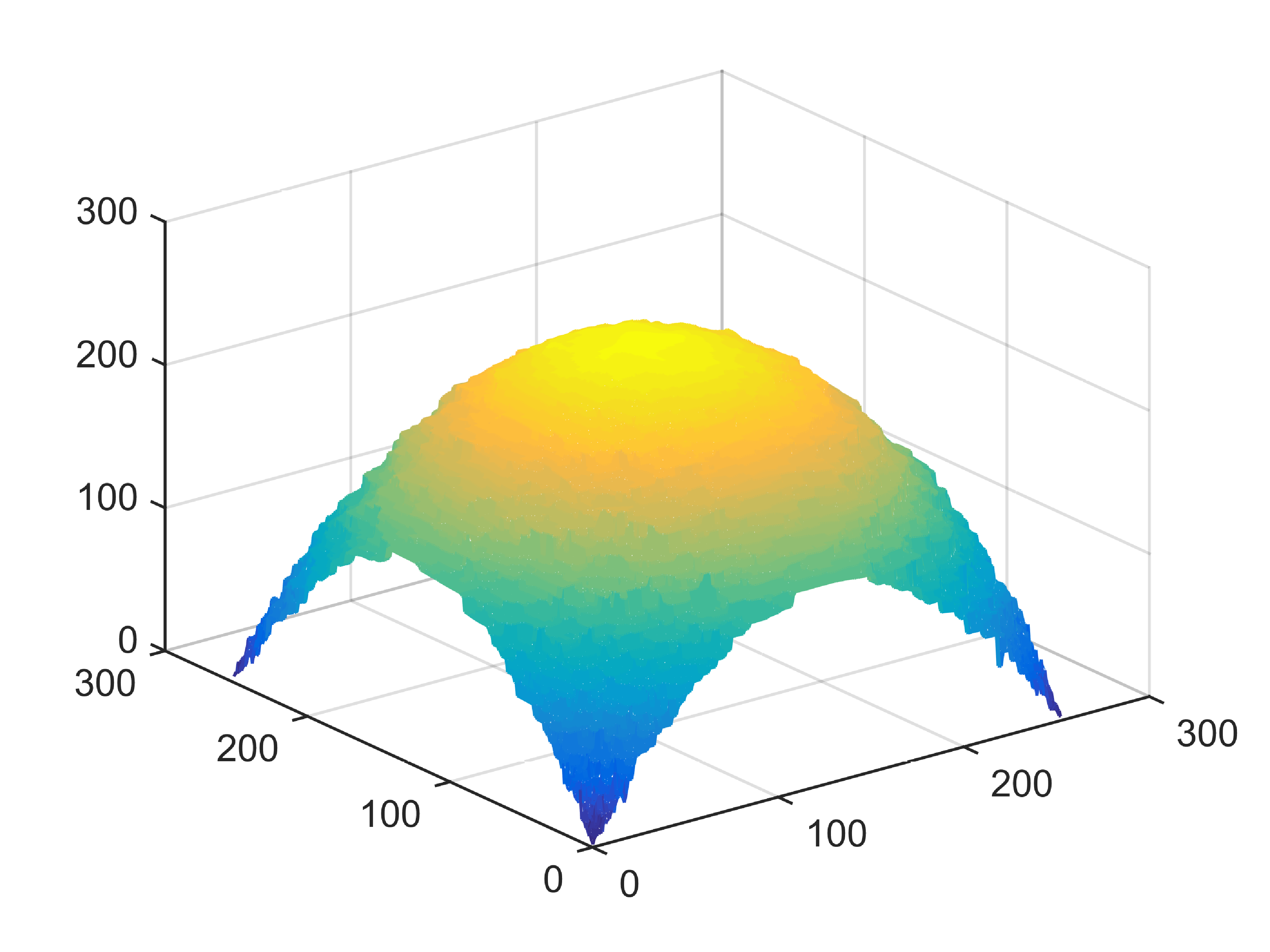}
		}
		\subfigure[AOS]{
			\includegraphics[width=0.19\textwidth]{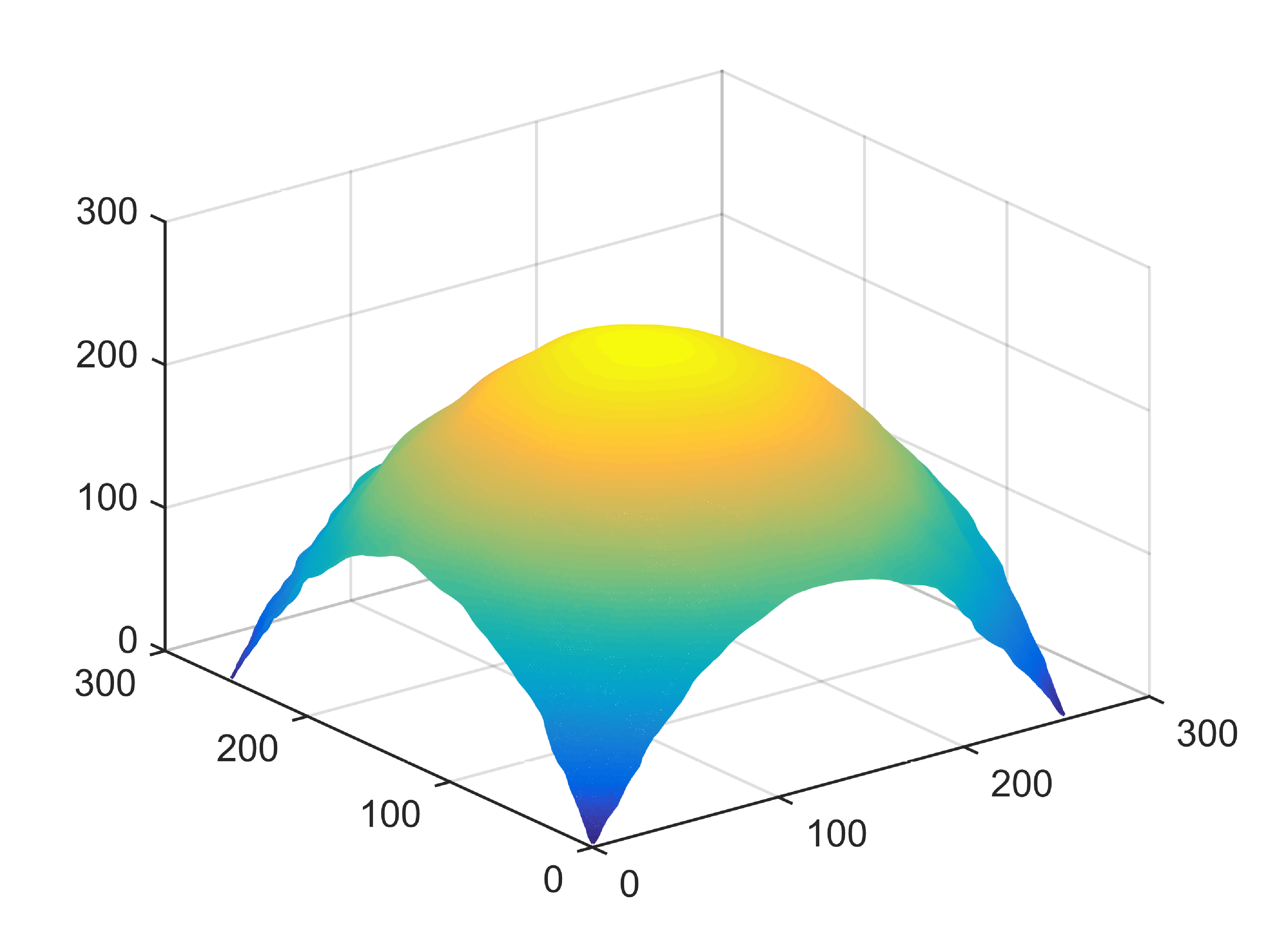}
		}
		\subfigure[SAV1]{
			\includegraphics[width=0.19\textwidth]{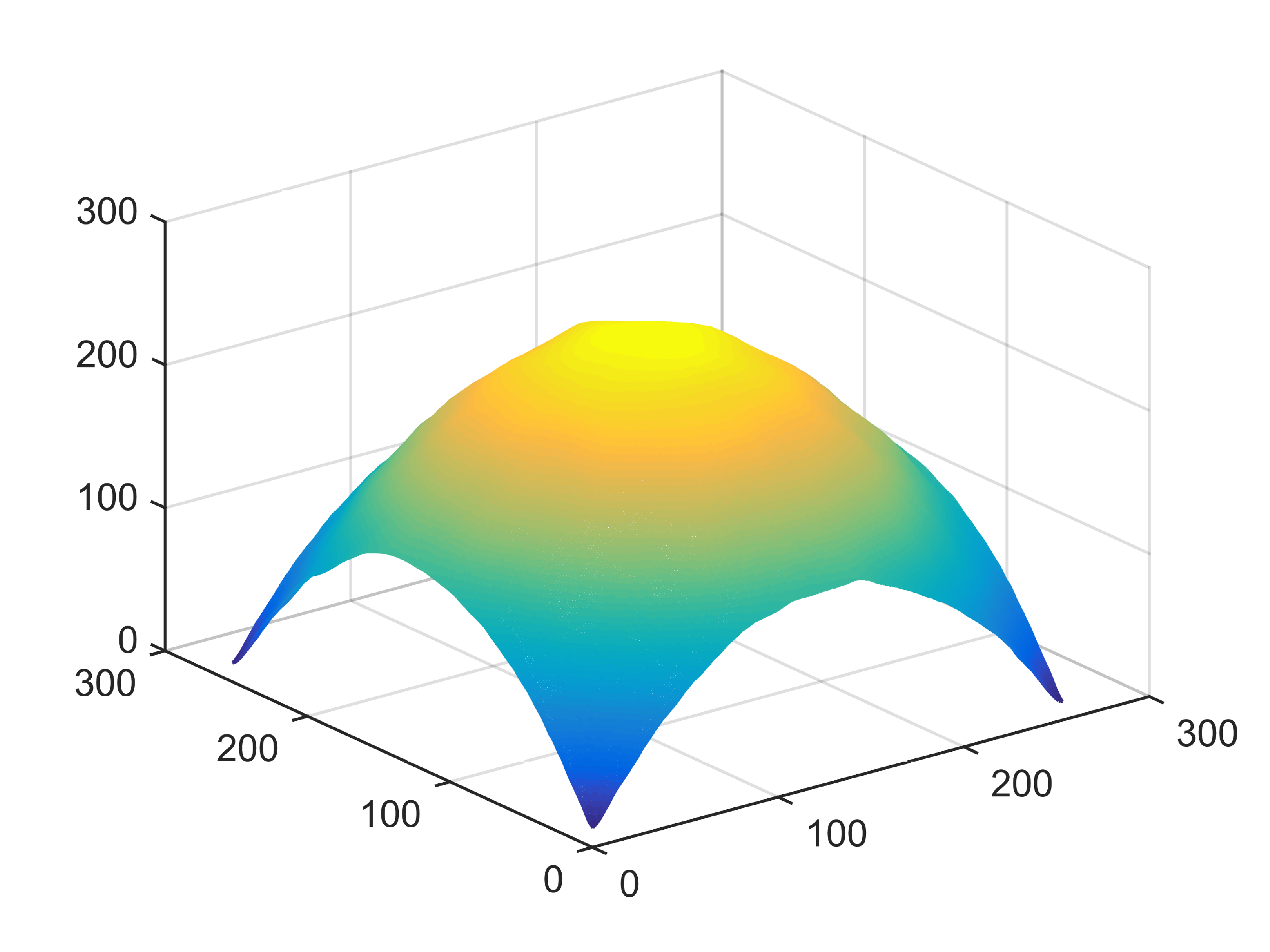}
		}
		\subfigure[SAV2]{
			\includegraphics[width=0.19\textwidth]{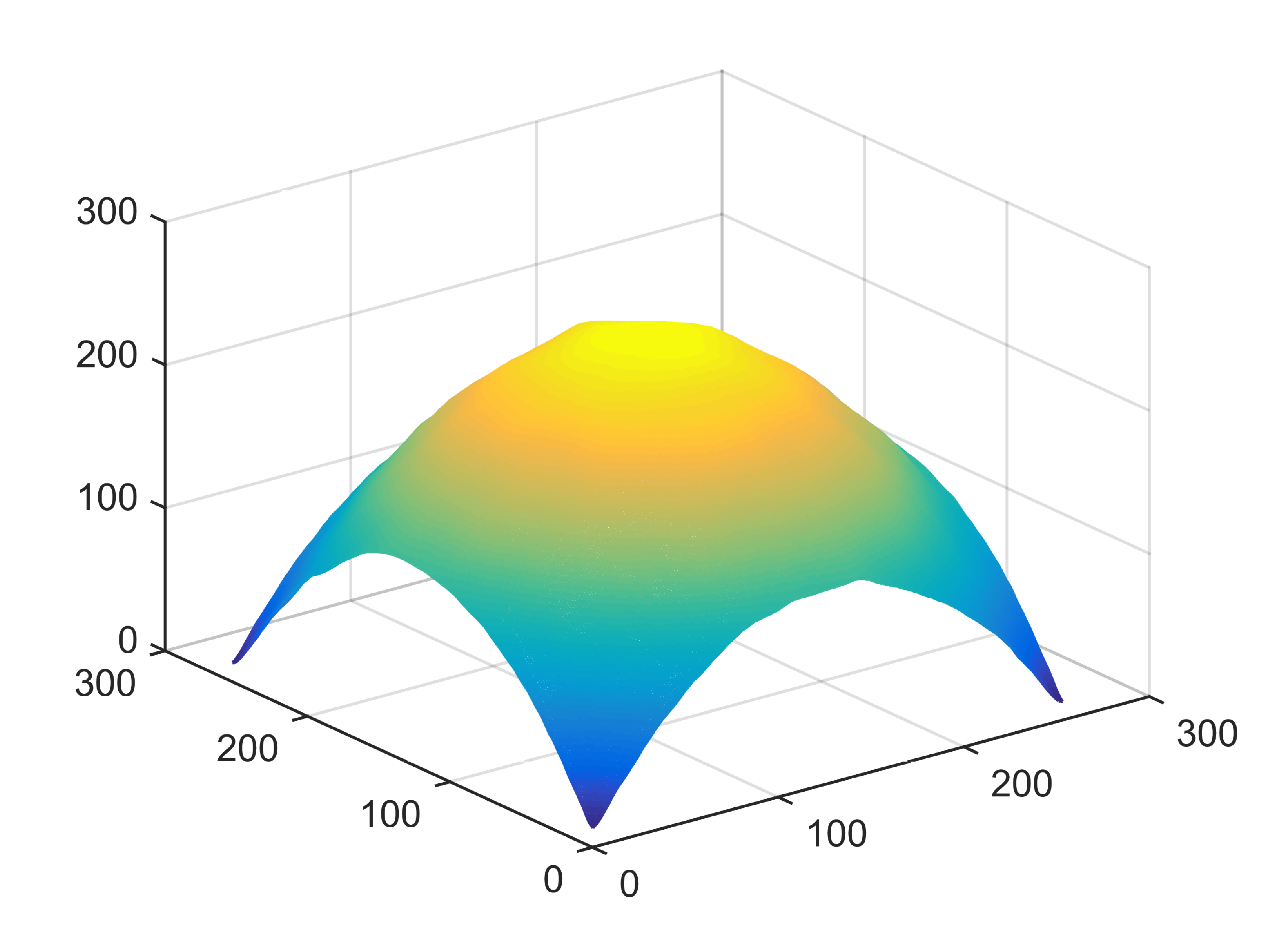}
		}
	\end{minipage}
	\caption{Surface plot of denoised Halo at noise level $L=10$}
	\label{mesh}
\end{figure}

In order to illustrate the superiority of the model and algorithm, we used gamma noise pollution with noise levels $L$ of $1$, $4$, and $10$ respectively. Generally, all models can suppress noise very well. However, the NTV model cannot effectively solve the high-noise situation of $L=1$, because the expectation and variance of the reciprocal of gamma noise are infinite when $L=1$. In order to quantitatively evaluate the model, we report the PSNR and SSIM values of the recovery results of various methods in Table \ref{tab:PSNRSSIM}, where the optimal experimental results are marked in bold. We usually choose a not too large weight $\beta$ for high-order terms to alleviate the step effect and assist the prior of the area term. On the one hand, we can see that in most cases the AOS algorithm can obtain experimental results quickly, but it also brings the effect of large errors, which can be found in the results of L=1 and L=4 for ``Halo'' in Table \ref{tab:PSNRSSIM}. Therefore, we actually recommend using the SAV algorithm to obtain higher quality experimental results.
\begin{figure}[h]
	\centering
	\begin{minipage}{\textwidth}
		\centering
		\subfigure[Noisy]{
			\includegraphics[width=0.23\textwidth]{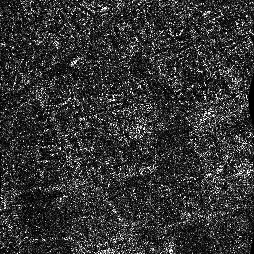}
		}
		\subfigure[AA]{
			\includegraphics[width=0.23\textwidth]{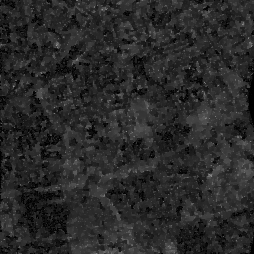}
		}
		\subfigure[DD]{
			\includegraphics[width=0.23\textwidth]{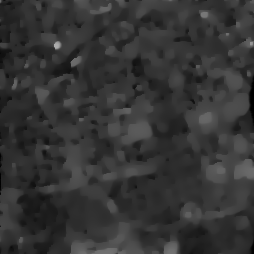}
		}
		\subfigure[MuLoG+BM3D]{
			\includegraphics[width=0.23\textwidth]{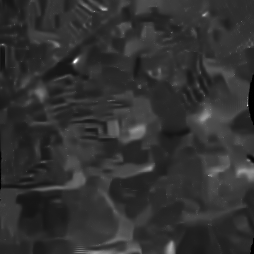}
		}
	\end{minipage}
	\begin{minipage}{\textwidth}
		\centering
		\subfigure[EE]{
			\includegraphics[width=0.23\textwidth]{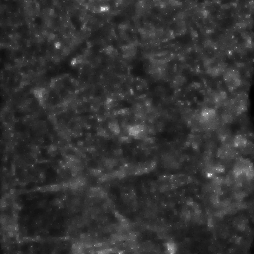}
		}
		\subfigure[AOS]{
			\includegraphics[width=0.23\textwidth]{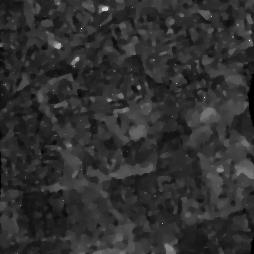}
		}
		\subfigure[SAV1]{
			\includegraphics[width=0.23\textwidth]{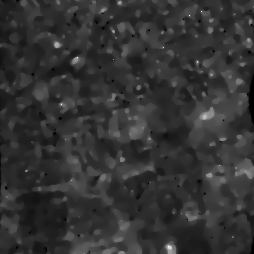}
		}
		\subfigure[SAV2]{
			\includegraphics[width=0.23\textwidth]{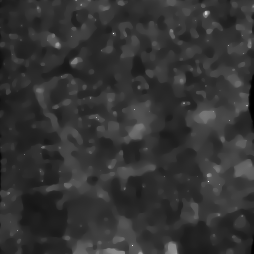}
		}
	\end{minipage}
	\caption{Denoised results of the Aerial. a Noisy:L=1. b-h:Denoised result}
	\label{aerialL1}
\end{figure}
\begin{figure}[h]
	\centering
	\begin{minipage}{\textwidth}
		\centering
		\subfigure[Noisy]{
			\includegraphics[width=0.23\textwidth]{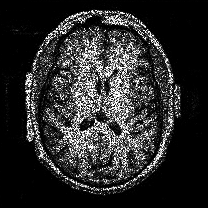}
		}
		\subfigure[AA]{
			\includegraphics[width=0.23\textwidth]{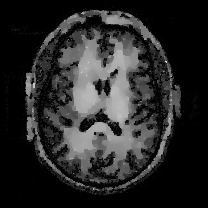}
		}
		\subfigure[DD]{
			\includegraphics[width=0.23\textwidth]{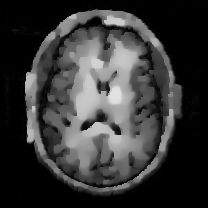}
		}
		\subfigure[MuLoG+BM3D]{
			\includegraphics[width=0.23\textwidth]{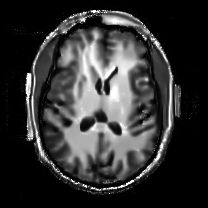}
		}
	\end{minipage}
	\begin{minipage}{\textwidth}
		\centering
		\subfigure[EE]{
			\includegraphics[width=0.23\textwidth]{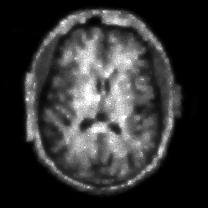}
		}
		\subfigure[AOS]{
			\includegraphics[width=0.23\textwidth]{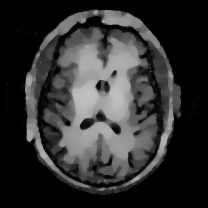}
		}
		\subfigure[SAV1]{
			\includegraphics[width=0.23\textwidth]{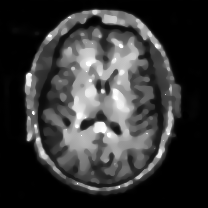}
		}
		\subfigure[SAV2]{
			\includegraphics[width=0.23\textwidth]{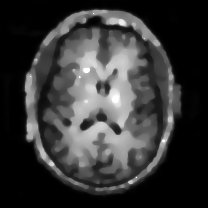}
		}
	\end{minipage}
	\caption{Denoised results of the Brain.a Noisy:L=1. b-h:Denoised result }
	\label{brainL1}
\end{figure}
\begin{figure}[h]
	\centering
	\begin{minipage}{\textwidth}
		\centering
		\subfigure[Noisy]{
			\includegraphics[width=0.23\textwidth]{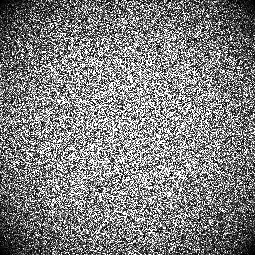}
		}
		\subfigure[AA]{
			\includegraphics[width=0.23\textwidth]{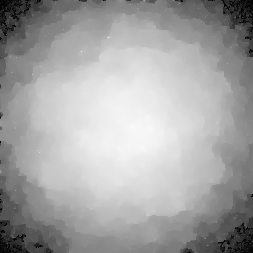}
		}
		\subfigure[DD]{
			\includegraphics[width=0.23\textwidth]{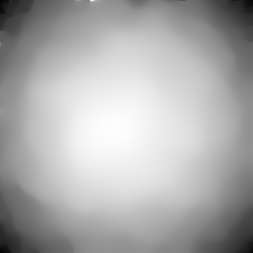}
		}
		\subfigure[MuLoG+BM3D]{
			\includegraphics[width=0.23\textwidth]{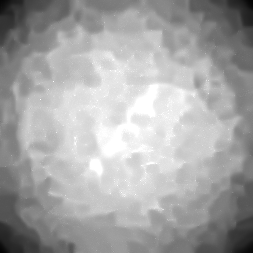}
		}
	\end{minipage}
	\begin{minipage}{\textwidth}
		\centering
		\subfigure[EE]{
			\includegraphics[width=0.23\textwidth]{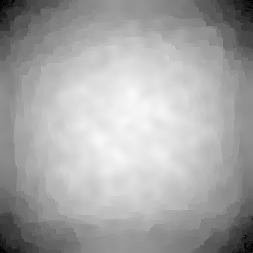}
		}
		\subfigure[AOS]{
			\includegraphics[width=0.23\textwidth]{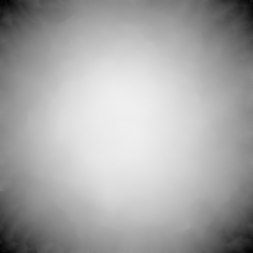}
		}
		\subfigure[SAV1]{
			\includegraphics[width=0.23\textwidth]{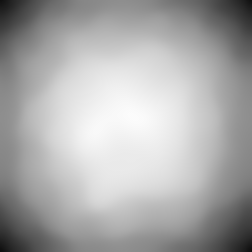}
		}
		\subfigure[SAV2]{
			\includegraphics[width=0.23\textwidth]{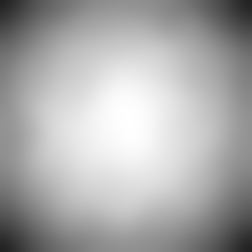}
		}
	\end{minipage}
	\caption{Denoising results of Halo at noise level $L=1$}
	\label{haloL1}
\end{figure}
\begin{figure}[h]
	\centering
	\begin{minipage}{\textwidth}
		\centering
		\subfigure[AA]{
			\includegraphics[width=0.23\textwidth]{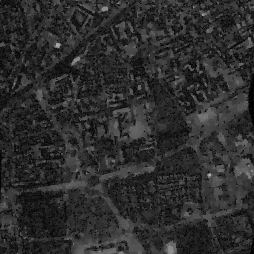}
		}
		\subfigure[DD]{
			\includegraphics[width=0.23\textwidth]{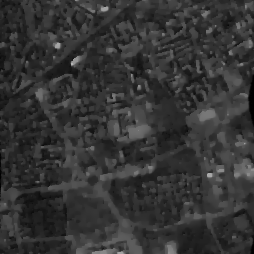}
		}
		\subfigure[MuLoG+BM3D]{
			\includegraphics[width=0.23\textwidth]{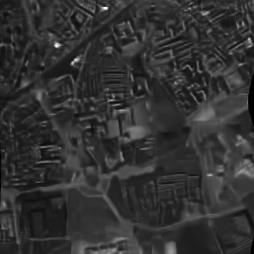}
		}
		\subfigure[NTV]{
			\includegraphics[width=0.23\textwidth]{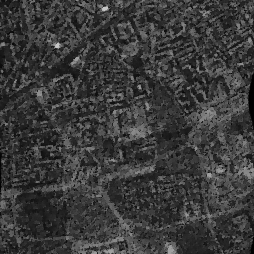}
		}
	\end{minipage}
	\begin{minipage}{\textwidth}
		\centering
		\subfigure[EE]{
			\includegraphics[width=0.23\textwidth]{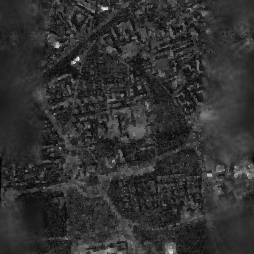}
		}
		\subfigure[AOS]{
			\includegraphics[width=0.23\textwidth]{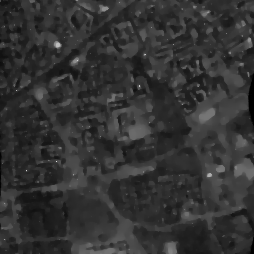}
		}
		\subfigure[SAV1]{
			\includegraphics[width=0.23\textwidth]{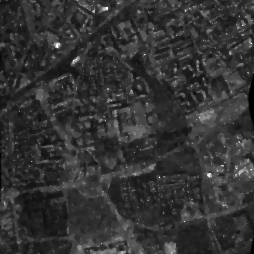}
		}
		\subfigure[SAV2]{
			\includegraphics[width=0.23\textwidth]{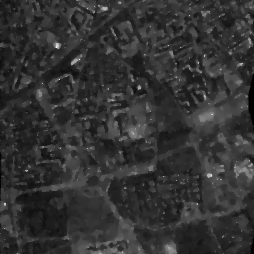}
		}
	\end{minipage}
	\caption{Denoising results of Aerial at noise level $L=4$}
	\label{aerialL4}
\end{figure}
\begin{figure}[h]
	\centering
	\begin{minipage}{\textwidth}
		\centering
		\subfigure[AA]{
			\includegraphics[width=0.23\textwidth]{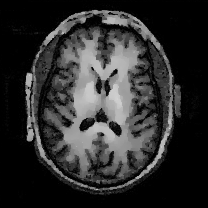}
		}
		\subfigure[DD]{
			\includegraphics[width=0.23\textwidth]{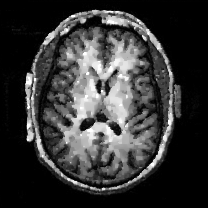}
		}
		\subfigure[MuLoG+BM3D]{
			\includegraphics[width=0.23\textwidth]{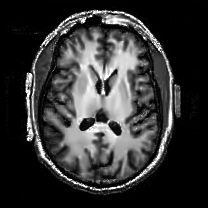}
		}
		\subfigure[NTV]{
			\includegraphics[width=0.23\textwidth]{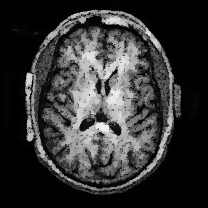}
		}
	\end{minipage}
	\begin{minipage}{\textwidth}
		\centering
		\subfigure[EE]{
			\includegraphics[width=0.23\textwidth]{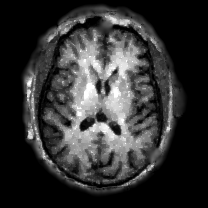}
		}
		\subfigure[AOS]{
			\includegraphics[width=0.23\textwidth]{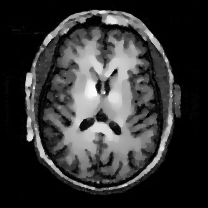}
		}
		\subfigure[SAV1]{
			\includegraphics[width=0.23\textwidth]{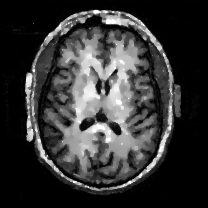}
		}
		\subfigure[SAV2]{
			\includegraphics[width=0.23\textwidth]{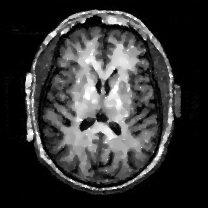}
		}
	\end{minipage}
	\caption{Denoising results of Brain at noise level $L=4$}
	\label{brainL4}
\end{figure}
\begin{figure}[h]
	\centering
	\begin{minipage}{\textwidth}
		\centering
		\subfigure[AA]{
			\includegraphics[width=0.23\textwidth]{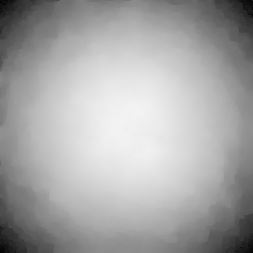}
		}
		\subfigure[DD]{
			\includegraphics[width=0.23\textwidth]{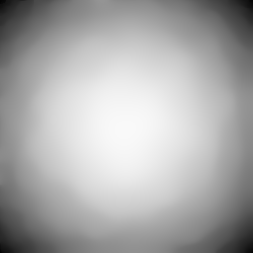}
		}
		\subfigure[MuLoG+BM3D]{
			\includegraphics[width=0.23\textwidth]{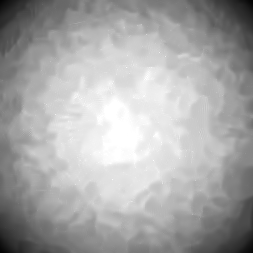}
		}
		\subfigure[NTV]{
			\includegraphics[width=0.23\textwidth]{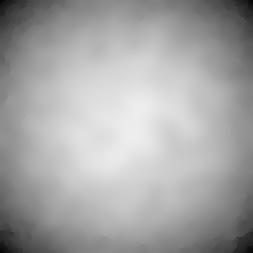}
		}
	\end{minipage}
	\begin{minipage}{\textwidth}
		\centering
		\subfigure[EE]{
			\includegraphics[width=0.23\textwidth]{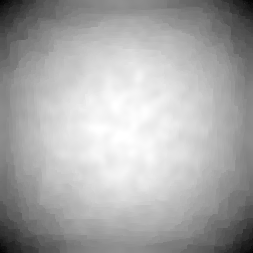}
		}
		\subfigure[AOS]{
			\includegraphics[width=0.23\textwidth]{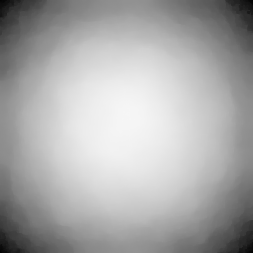}
		}
		\subfigure[SAV1]{
			\includegraphics[width=0.23\textwidth]{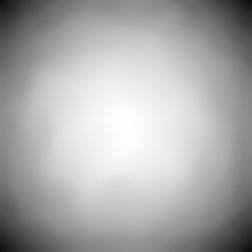}
		}
		\subfigure[SAV2]{
			\includegraphics[width=0.23\textwidth]{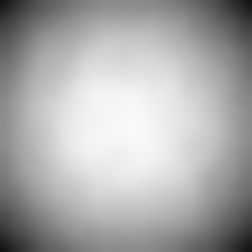}
		}
	\end{minipage}
	\caption{Denoising results of Halo at noise level $L=4$}
	\label{haloL4}
\end{figure}
\begin{figure}[h]
	\centering
	\begin{minipage}{\textwidth}
		\centering
		\subfigure[AA]{
			\includegraphics[width=0.23\textwidth]{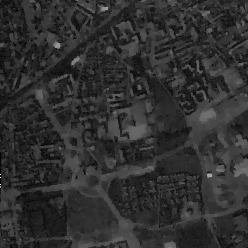}
		}
		\subfigure[DD]{
			\includegraphics[width=0.23\textwidth]{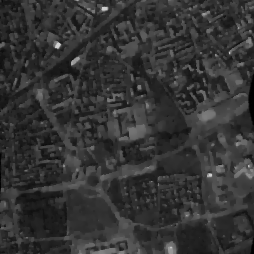}
		}
		\subfigure[MuLoG+BM3D]{
			\includegraphics[width=0.23\textwidth]{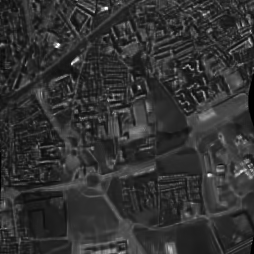}
		}
		\subfigure[NTV]{
			\includegraphics[width=0.23\textwidth]{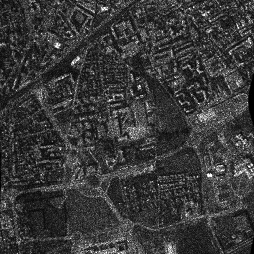}
		}
	\end{minipage}
	\begin{minipage}{\textwidth}
		\centering
		\subfigure[EE]{
			\includegraphics[width=0.23\textwidth]{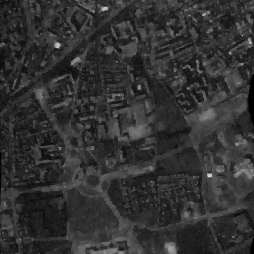}
		}
		\subfigure[AOS]{
			\includegraphics[width=0.23\textwidth]{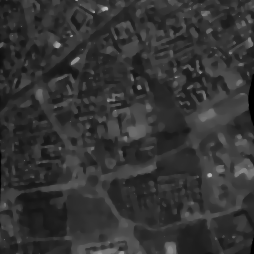}
		}
		\subfigure[SAV1]{
			\includegraphics[width=0.23\textwidth]{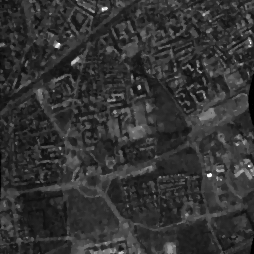}
		}
		\subfigure[SAV2]{
			\includegraphics[width=0.23\textwidth]{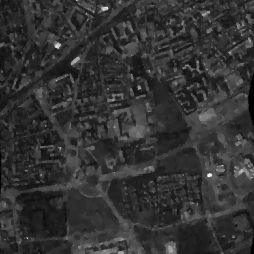}
		}
	\end{minipage}
	\caption{Denoising results of Aerial at noise level $L=10$}
	\label{aerialL10}
\end{figure}
\begin{figure}[h]
	\centering
	\begin{minipage}{\textwidth}
		\centering
		\subfigure[AA]{
			\includegraphics[width=0.23\textwidth]{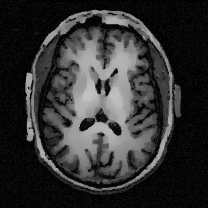}
		}
		\subfigure[DD]{
			\includegraphics[width=0.23\textwidth]{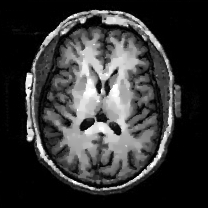}
		}
		\subfigure[MuLoG+BM3D]{
			\includegraphics[width=0.23\textwidth]{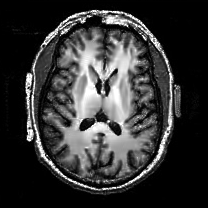}
		}
		\subfigure[NTV]{
			\includegraphics[width=0.23\textwidth]{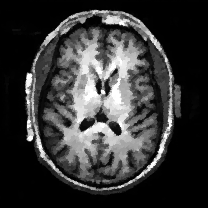}
		}
	\end{minipage}
	\begin{minipage}{\textwidth}
		\centering
		\subfigure[EE]{
			\includegraphics[width=0.23\textwidth]{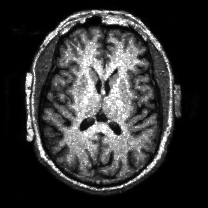}
		}
		\subfigure[AOS]{
			\includegraphics[width=0.23\textwidth]{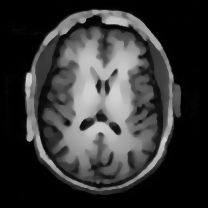}
		}
		\subfigure[SAV1]{
			\includegraphics[width=0.23\textwidth]{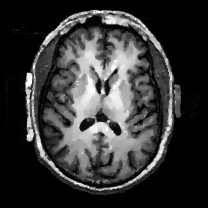}
		}
		\subfigure[SAV2]{
			\includegraphics[width=0.23\textwidth]{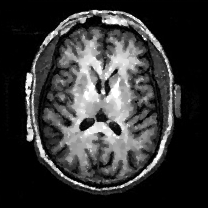}
		}
	\end{minipage}
	\caption{Denoising results of Brain at noise level $L=10$}
	\label{brainL10}
\end{figure}
\begin{figure}[h]
	\centering
	\begin{minipage}{\textwidth}
		\centering
		\subfigure[AA]{
			\includegraphics[width=0.23\textwidth]{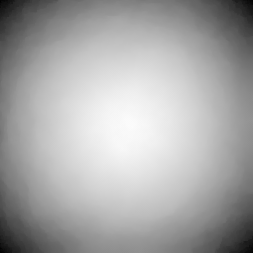}
		}
		\subfigure[DD]{
			\includegraphics[width=0.23\textwidth]{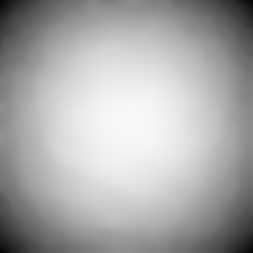}
		}
		\subfigure[MuLoG+BM3D]{
			\includegraphics[width=0.23\textwidth]{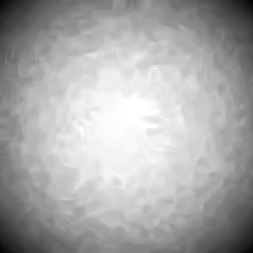}
		}
		\subfigure[NTV]{
			\includegraphics[width=0.23\textwidth]{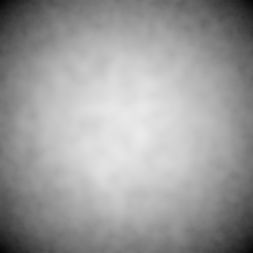}
		}
	\end{minipage}
	\begin{minipage}{\textwidth}
		\centering
		\subfigure[EE]{
			\includegraphics[width=0.23\textwidth]{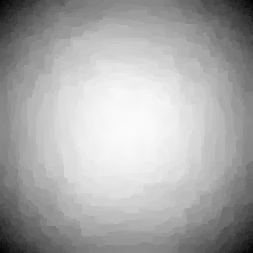}
		}
		\subfigure[AOS]{
			\includegraphics[width=0.23\textwidth]{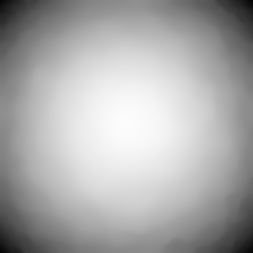}
		}
		\subfigure[SAV1]{
			\includegraphics[width=0.23\textwidth]{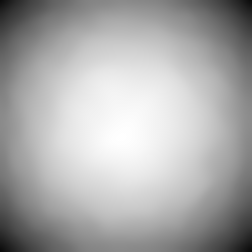}
		}
		\subfigure[SAV2]{
			\includegraphics[width=0.23\textwidth]{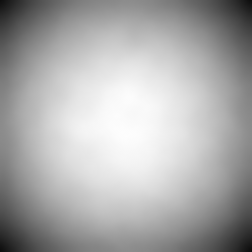}
		}
	\end{minipage}
	\caption{Denoising results of Halo at noise level $L=10$}
	\label{haloL10}
\end{figure}
\begin{figure}[h]
	\centering
	\begin{minipage}{\textwidth}
		\centering
		\subfigure[AA]{
			\includegraphics[width=0.23\textwidth]{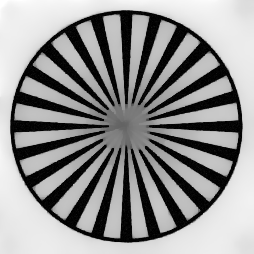}
		}
		\subfigure[DD]{
			\includegraphics[width=0.23\textwidth]{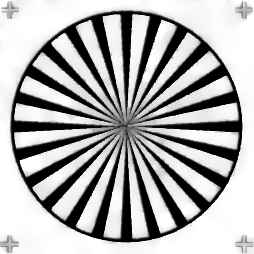}
		}
		\subfigure[MuLoG+BM3D]{
			\includegraphics[width=0.23\textwidth]{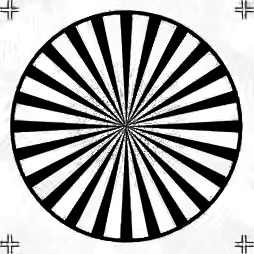}
		}
		\subfigure[NTV]{
			\includegraphics[width=0.23\textwidth]{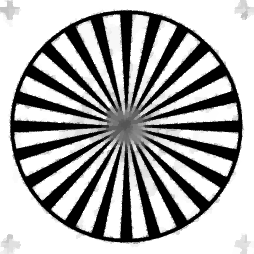}
		}
	\end{minipage}
	\begin{minipage}{\textwidth}
		\centering
		\subfigure[EE]{
			\includegraphics[width=0.23\textwidth]{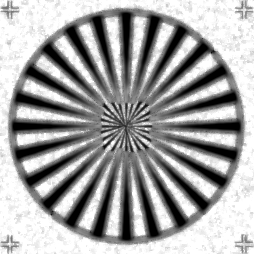}
		}
		\subfigure[AOS]{
			\includegraphics[width=0.23\textwidth]{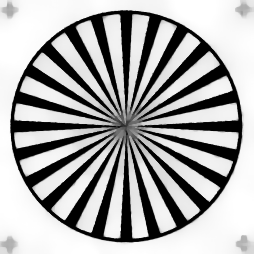}
		}
		\subfigure[SAV1]{
			\includegraphics[width=0.23\textwidth]{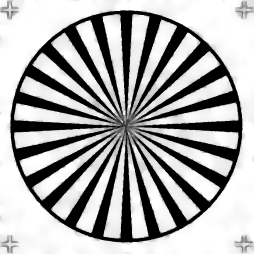}
		}
		\subfigure[SAV2]{
			\includegraphics[width=0.23\textwidth]{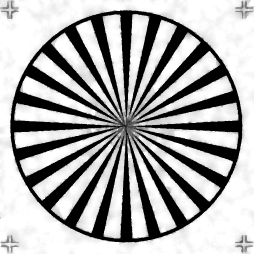}
		}
	\end{minipage}
	\caption{Denoising results of Dartboard at noise level $L=10$}
	\label{dartboardL10}
\end{figure}
When $L=1$, an obvious feature is that quite a few white point outliers appear randomly in the image. And an advantage of the diffusion equation-based method is that it will smooth the image at any noise level. In contrast, these random outliers appearing in MuLoG's denoising results brings bad visual effects. These phenomena are depicted in Fig. \ref{haloL1}. In addition, DD, EE and our model based on adaptive area denoising can restore various areas of the image to varying degrees, and can well protect the real texture of the image. Meanwhile, the MuLoG method can effectively protect the contrast and tiny texture of the image.
From Fig. \ref{brainL1}, \ref{brainL4}, \ref{brainL10}, we can clearly see that the AA model exposes so relatively large linear region that is prone to occur in low-order models. Although NTV can remove noise very well, due to the limitations of the model itself, it is difficult to remove large noise, and there are still dirty spots in Fig. \ref{brainL4} that cannot be removed. It is worth pointing out that due to the use of block matching technology, the MuLoG method often appears artificial boundaries, which can be easily seen from the center of Dartboard in \ref{dartboardL10}. Inevitably, in addition, for ``Halo'', this method fails greatly, the false boundaries almost occupy the entire image, and it is unbearable. Compared with the DD and EE models, the model we proposed can protect the boundaries very well and obtain a denoising effect with richer details. This can be seen further when we plot the surface figure of ``Halo'' in Fig. \ref{mesh}. From Fig. \ref{dartboardL10} we can see that our model has the best performance among all models in maintaining the smoothness and connectivity of image lines, which benefits from the curvature model's prior regularization of image curves. 
As can be seen from Table \ref{tab:PSNRSSIM}, although MuLoG has more artificial boundaries, it still achieves best PSNR and SSIM in some figures. Compared with other methods, our model can effectively avoid the staircasing effects on natural images, SAR images, ultrasound images, and synthetic images. It can effectively protect detailed textures and smooth noise. Especially for ``Halo'', which is a synthetic images with large slopes, the model we proposed has achieved amazing experimental results and can almost restore the original image, far exceeding other models. Therefore, our model is very suitable for the denoising process of gradient images.
\begin{table}[]
	\caption{PSNR and SSIM between various multiplicative denoising methods on test images corrupted by Gamma noise}
	\label{tab:PSNRSSIM}       
	\begin{tabular}{llllllll}
		\hline\noalign{\smallskip}
		Image & Model      &       & PSNR  &       &      & SSIM &      \\ \noalign{\smallskip}\hline\noalign{\smallskip}
		& L          & 1     & 4     & 10    & 1    & 4    & 10   \\ 	\noalign{\smallskip}\hline
		& AA         & 21.66 & 23.76 & 19.86 & 0.45 & 0.65 & 0.57 \\
		& DD         & 22.39 & 24.94 & 25.97 & 0.48 & 0.68 & $\bm{0.80}$ \\
		& MuLoG+BM3D & 22.18 & 24.86 & $\bm{27.30}$ & 0.40 & 0.65 & $\bm{0.80}$ \\
		Aerial	& NTV        & -     & 23.43 & 24.07 & -    & 0.66 & 0.64 \\
		& EE         & $\bm{22.62}$ & 23.45 & 26.79 & $\bm{0.51}$ & 0.59 & 0.78 \\
		& AOS        & 22.52 & 24.85 & 26.92 & 0.50 & 0.68 & 0.79 \\
		& SAV1       & 22.54 & $\bm{24.92}$ & 27.11 & 0.50 & $\bm{0.69}$ & $\bm{0.80}$ \\
		& SAV2       & 22.53 & 24.88 & 27.02 & 0.49 & 0.68 & $\bm{0.80}$ \\ 
		& AA         & 16.83 & 19.75 & 20.32 & 0.68 & 0.81 & 0.57 \\
		& DD         & 18.93 & 22.01 & 24.39 & 0.74 & 0.85 & 0.90 \\
		& MuLoG+BM3D & $\bm{19.51}$ & 22.47 & 24.50 & $\bm{0.78}$ & $\bm{0.87}$ & $\bm{0.91}$ \\
		Brain	& NTV        & -     & 21.22 & 23.65 & -    & 0.83 & 0.87 \\
		& EE         & 18.93 & 21.04 & 23.24 & 0.75 & 0.84 & 0.89 \\
		& AOS        & 19.23 & 22.12 & 24.10 & 0.75 & 0.84 & 0.88 \\
		& SAV1       & 19.36 & $\bm{22.55}$ & 24.73 & 0.74 & 0.86 & 0.90 \\
		& SAV2       & 19.39 & 22.50 & $\bm{24.75}$ & 0.75 & 0.86 & 0.90 \\ 
		& AA         & 27.86 & 31.50 & 33.31 & 0.94 & 0.98 & 0.99 \\
		& DD         & 33.00 & 36.66 & 39.05 & $\bm{0.99}$ & $\bm{0.99}$ & $\bm{1.00}$ \\
		& MuLoG+BM3D & 26.20 & 30.79 & 33.31 & 0.78 & 0.87 & 0.88 \\
		Halo	& NTV        & -     & 27.22 & 27.88 & -    & 0.98 & 0.97 \\
		& EE         & 29.89 & 34.93 & 36.74 & 0.92 & 0.94 & 0.94 \\
		& AOS        & 32.13 & 37.18 & $\bm{40.66}$ & $\bm{0.99}$ & $\bm{0.99}$ & $\bm{1.00}$ \\
		& SAV1       & $\bm{33.97}$ & $\bm{37.75}$ & 40.60 & $\bm{0.99}$ & $\bm{0.99}$ & 0.99 \\
		& SAV2       & $\bm{33.97}$ & 37.58 & 40.63 & 0.96 & $\bm{0.99}$ & 0.99 \\ 
		\noalign{\smallskip}\hline
	\end{tabular}
\end{table}

\subsection{Discussion between First-order and Second-order SAV}
\label{sec:4subsec:5}
\begin{figure}[h]
	\centering
	\begin{minipage}{\textwidth}
		\centering
		\subfigure[PSNR]{
			\includegraphics[width=0.48\textwidth]{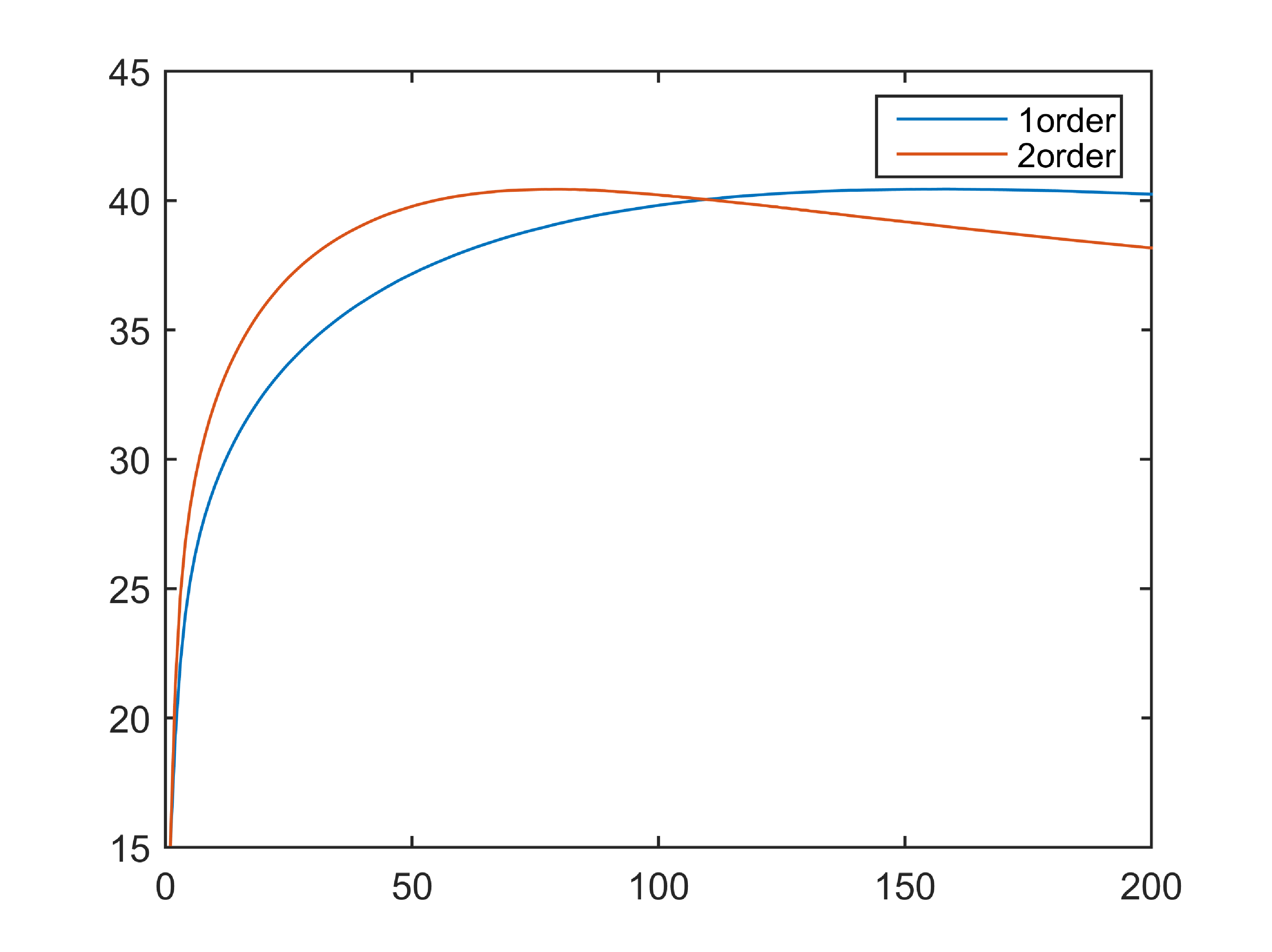}
		}
		\subfigure[Energy]{
			\includegraphics[width=0.48\textwidth]{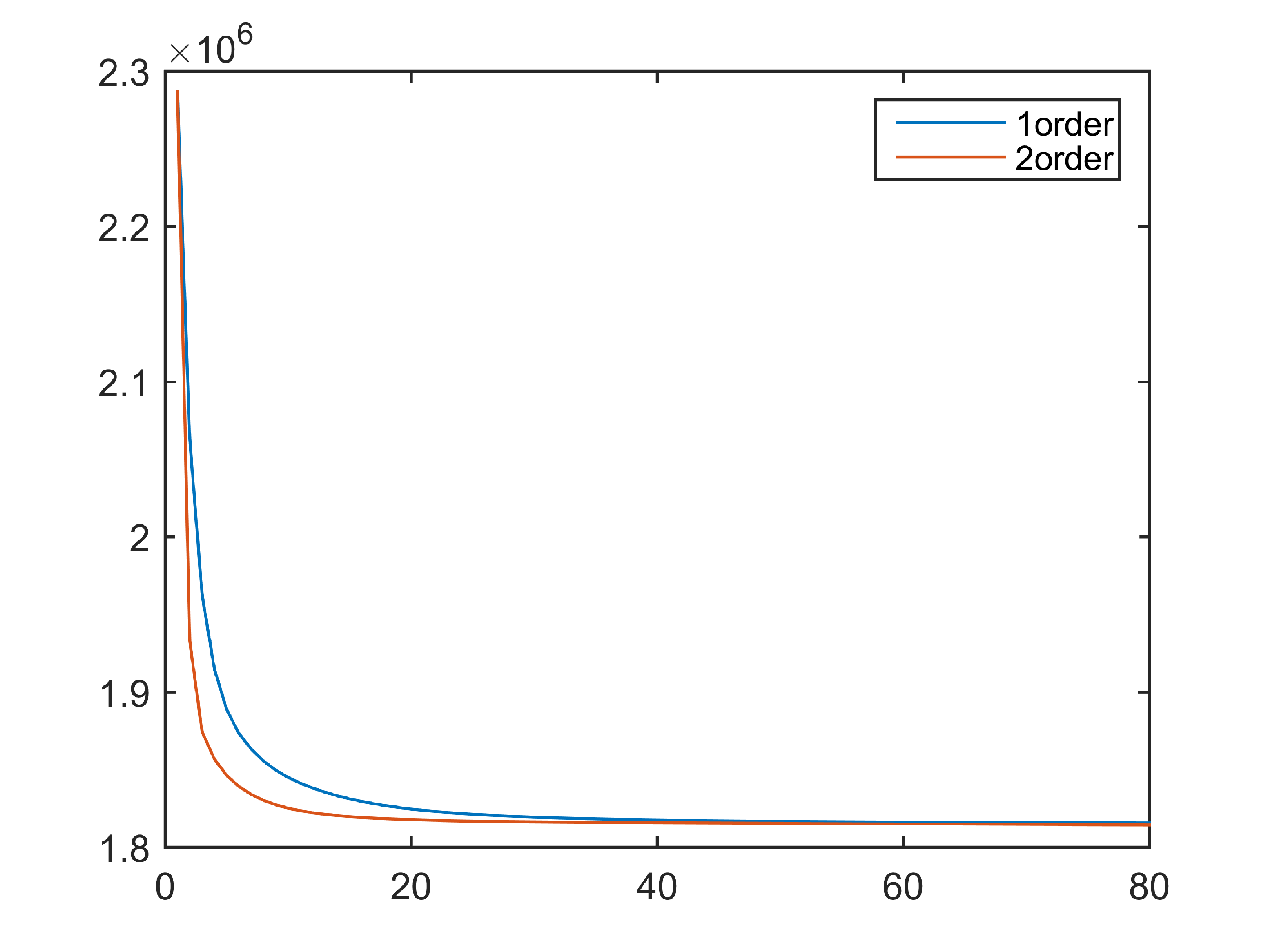}
		}
	\end{minipage}
	\caption{Denoising results of Hole at noise level $L=10$. We fix $b=0.0001$, Constant=$10^8$, $T=200$. We set $\tau_{min}=1.8$, $\tau_{max}=2$ in the second-order scheme and $\tau_{min}=0.8$, $\tau_{max}=1$ in the first-order scheme}
	\label{12orderComparison}
\end{figure}

\begin{figure}[h]
	\centering
	\begin{minipage}{\textwidth}
		\centering
		\subfigure[PSNR]{
			\includegraphics[width=0.48\textwidth]{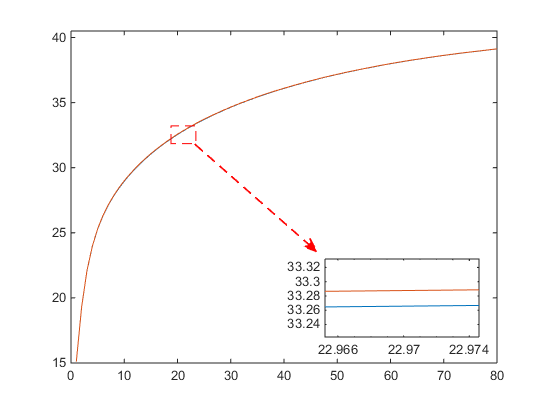}
		}
		\subfigure[Energy]{
			\includegraphics[width=0.48\textwidth]{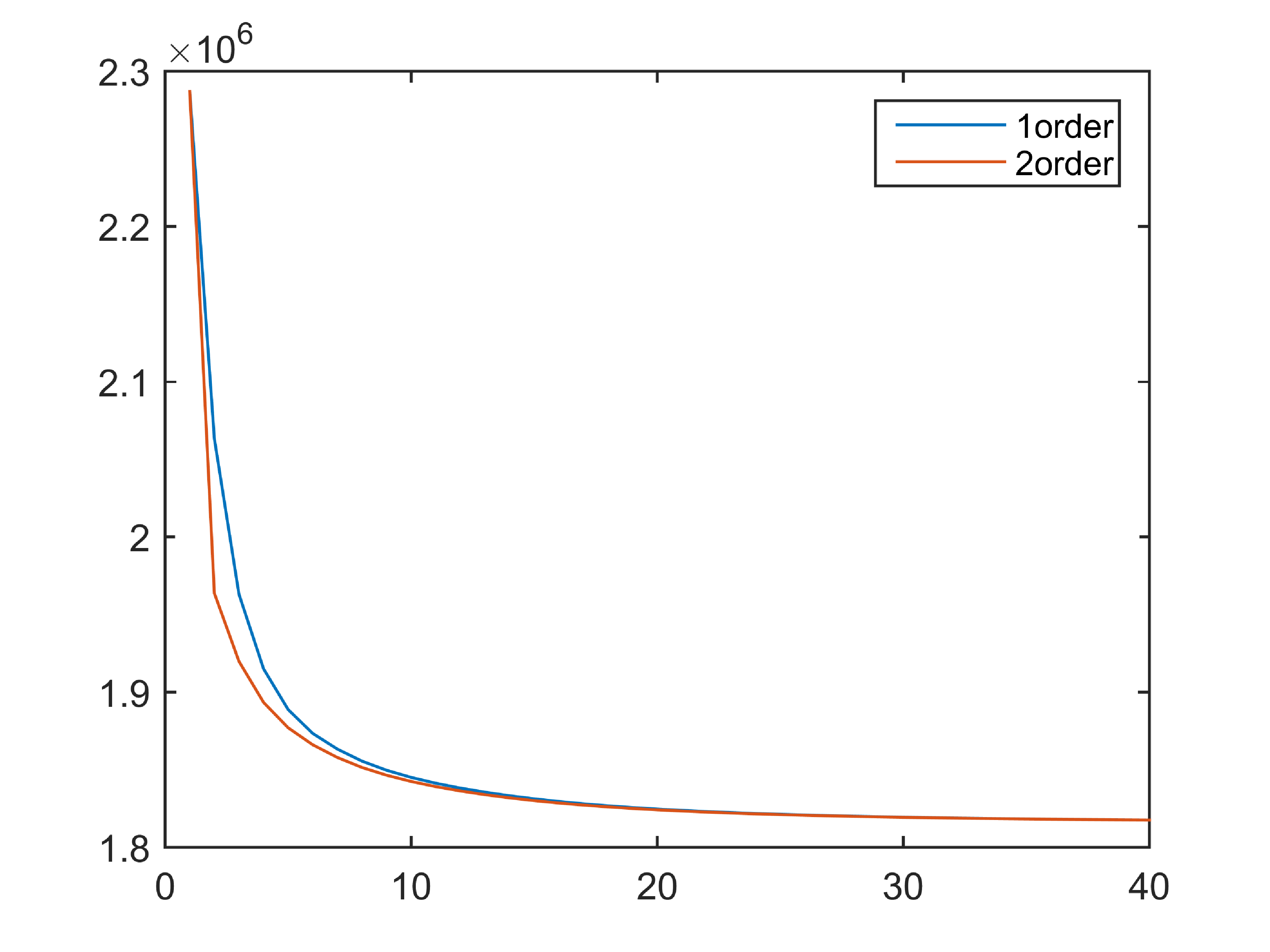}
		}
	\end{minipage}
	\caption{Denoising results of Hole at noise level $L=10$. We fix $b=0.0001$, Constant=$10^8$, $T=200$, $\tau_{min}=1.8$, $\tau_{max}=2$ for both sheme}
	\label{12orderComparison_Samestep}
\end{figure}

In order to better verify the advantages of the second-order SAV method, the PSNR and energy drop curves of each order model in the image ``Hole'' are shown in Fig. \ref{12orderComparison}. We set the remaining parameters keep equal except for the time step size in the model. Compared with first-order SAV, we can clearly see that the second-order model can achieve the best denoising effect when iterating for 80 steps, while the first-order model requires 200 iterations to achieve the same denoising effect. The second-order SAV algorithm reaches the best experimental results faster and decay rapidly. Fig. \ref{12orderComparison}b illustrates that the higher-order model can perform energy reduction faster. This means that in the second-order scheme, we can use a larger step size and achieve the same error accuracy, which is consistent with the theoretical error analysis of the numerical scheme. In order to further eliminate the iteration acceleration caused by the step size factor, we choose to choose the same time step for the two schemes. As shown in Fig. \ref{12orderComparison_Samestep}b, we can find that the second-order algorithm still performs well under the same step size. Compared with the first-order algorithm, it has the advantage in terms of computing speed. In addition, due to the advantages of the model itself, the convergence of PSNR in the two algorithms is almost the same, but after zooming in, it can be found that the second-order algorithm always keep it slightly higher with a certain gap. Under the premise of the same step size, the second-order algorithm can improve the denoising effect to a certain extent compared with the first-order algorithm.


\section{Conclusions}
\label{sec:5}
In this paper, we propose a mixed geometry information model to deal with the multiplicative denoising task. The model we propose can well match the characteristics. Based on the surface area and curvature, we designed the mixed geometry information term to protect the edges and texture of the image. At the same time, in order to efficiently address the difficulties caused by solving highly nonlinear models, we proposed AOS and SAV algorithms to achieve unconditional stability. The SAV algorithm is more recommended to solve our model due to its higher accuracy. Numerical experiments are implemented on SAR images, ultrasound images and synthetic images, and various experimental results demonstrate the superiority and excellent performance of our model. Compared with other high-order methods, our model can well maintain image contrast, protect detailed texture and edge information. In addition, quantitative analysis results show that our proposed model has ideal experimental results compared with other state-of-the-art multiplicative denoising models.

\section*{Data availability statement}
No new data were created or analysed in this study.
\section*{Acknowledgements}
The work is partially supported by the National Natural Science Foundation of China (Nos. U21B2075, 12171123, 12101158 and 12301536), the Fundamental Research Funds for the Central Universities grant (Nos. 2022FRFK060020 and 2022FRFK060029), the Natural Sciences Foundation of Heilongjiang Province (No. ZD2022A001).
\section*{References}
\bibliographystyle{unsrt} 
\bibliography{Reference.bib}
\end{document}